\newif\ifSciRob
\SciRobfalse

\ifSciRob\else
\documentclass[9pt,twocolumn,twoside]{osajnl} 
\title{Bicycle Acrobatics with Reinforcement Learning}
\author[1$\dagger$$\ast$]{Shamel Fahmi}
\author[1$\dagger$]{Arianna Ilvonen}
\author[1$\dagger$]{Samuel Zapolsky}
\author[1$\dagger$]{Yu-Ming Chen}
\author[1$\ddagger$]{Ravi Boggavarapu}
\author[1$\ddagger$]{Paul Drews}
\author[1$\ddagger$]{Ashwin Khadke}
\author[1$\ddagger$]{Dean Molinaro}
\author[1$\ddagger$]{Kaiyu Zheng}
\author[1$\ddagger\ddagger$]{Alfred Rizzi}
\author[1$\ddagger\ddagger$]{Gabriel Nelson}
\affil[1]{Robotics and AI Institute (RAI), Cambridge, MA, USA.}
\affil[$\dagger$]{Core contributors listed in order of relative contribution.}
\affil[$\ddagger$]{Additional contributors listed alphabetically.}
\affil[$\ddagger\ddagger$]{Project leads listed in reverse order of relative contribution.}
\affil[$\ast$]{Corresponding author: sfahmi@rai-inst.com}
\newcommand*{\journalname}{Robotics and AI Institute}
\newcommand*{\journalshorttype}{Letter}
\newcommand*{\journallongtype}{Research Article - Preprint}
\definecolor{color2}{RGB}{0,51,153} 
\definecolor{color2b}{RGB}{0,51,153} 
\fi

\ifSciRob
\documentclass[12pt]{article} 
\usepackage{newtxtext,newtxmath}
\usepackage{scicite}
\usepackage{float}
\usepackage{url}
\usepackage{stfloats}
\usepackage{graphicx}
\usepackage{cuted}
\usepackage{caption}
\usepackage{newfloat}
\usepackage[table]{xcolor}
\usepackage[letterpaper,margin=1in]{geometry}
\renewenvironment{abstract}{\quotation}{\endquotation} 
\date{} 
\makeatletter
\renewcommand{\fnum@figure}{\textbf{Figure \thefigure}}
\renewcommand{\fnum@table}{\textbf{Table \thetable}}
\makeatother
\def\scititle{Bicycle Acrobatics with Reinforcement Learning}
\title{\bfseries \boldmath \scititle}
\author{%
Shamel Fahmi$^{1\dagger\ast}$, 
Arianna Ilvonen$^{1\dagger}$, 
Samuel Zapolsky$^{1\dagger}$, 
Yu-Ming Chen$^{1\dagger}$,\\
Ravi Boggavarapu$^{1\ddagger}$, 
Paul Drews$^{1\ddagger}$, 
Ashwin Khadke$^{1\ddagger}$, 
Dean Molinaro$^{1\ddagger}$,\\ 
Kaiyu Zheng$^{1\ddagger}$, 
Alfred Rizzi$^{1\ddagger\ddagger}$, and
Gabriel Nelson$^{1\ddagger\ddagger}$\\
\small$^{1}$RAI Institute, Cambridge MA, USA.\\
\small$^{\dagger}$Core contributors listed in order of relative contribution.\\
\small$^{\ddagger}$Additional contributors listed alphabetically.\\
\small$^{\ddagger\ddagger}$Project leads listed in reverse order of relative contribution.\\
\small$^\ast$Corresponding author. Email: \texttt{sfahmi@rai-inst.com}
}
\fi

\usepackage[]{units} 
\usepackage{xspace} 
\usepackage{multirow, booktabs, arydshln}
\usepackage{colortbl}
\newcommand{\ourrow}{\rowcolor{gray!20}}

\newcolumntype{x}[1]{>{\centering\arraybackslash\hspace{0pt}}p{#1}}
\usepackage{bbding}  
\usepackage{pifont}  
\usepackage[nonumberlist, nogroupskip, section=section, numberedsection=autolabel]{glossaries} 
\newacronym{rl}{RL}{Reinforcement Learning}
\newacronym{umv}{UMV}{Ultra Mobility Vehicle}
\newacronym{cmdp}{CMDP}{Constrained Markov Decision Process}
\newacronym{mdp}{MDP}{Markov Decision Process}
\newacronym{rsi}{RSI}{Reference State Initialization}
\newacronym{ppo}{PPO}{Proximal Policy Optimization}
\newacronym{mlp}{MLPs}{Multi-Layer Perceptrons}
\newacronym{rlhf}{RLHF}{Reinforcement Learning from Human Feedback}
\newacronym{dofs}{DOFs}{Degrees of Freedom}
\newacronym{imi}{IMI}{Iterative Motion Imitation}
\newacronym{fsm}{FSM}{Finite State Machine}
\newacronym{bt}{BT}{Behavior Tree}
\newacronym{mocap}{MoCap}{Motion Capture}
\newacronym{imu}{IMU}{Inertial Measurement Unit}
\makeglossaries
\glsdisablehyper 
\newcommand{\ie}{i.e.,\xspace}
\newcommand{\etal}{et~al.,\xspace}
\newcommand{\Rnum}{\mathbb{R}} 
\newcommand{\fref}[1]{Figure~\ref{#1}} 
\newcommand{\sref}[1]{Section~\ref{#1}} 
\newcommand{\tref}[1]{Table~\ref{#1}} 
\newcommand{\rev}[1]{}
\begin{document} 
\begin{abstract} \bfseries \boldmath \sffamily
Bicycle robots are fast and energy efficient, 
but their simple mechanical design and their underactuated and non-holonomic dynamics
make highly agile maneuvers difficult to achieve. 
Here, we use Reinforcement Learning (RL) to enable a bicycle robot to 
learn and compose a diverse repertoire of dynamic acrobatic stunts. 
Using 
different RL formulations such as
waypoint following, pose reaching, twist tracking, guided tracking, and motion imitation, 
the robot acquires 
autonomous single and multi-table forward and lateral jumps, 
steerable jumps, 
front flips, 
kip-ups, kip-downs, 
driving, wheelies, 
bunny hops, and three-point turns. 
To coordinate these behaviors, 
we introduce an orchestrator that transitions 
between policies using state-dependent triggers, 
enabling robust long-horizon acrobatic stunts. 
We validate the approach on the Ultra Mobility Vehicle (UMV), 
a custom bicycle robot, 
in simulation and hardware. 
The robot 
repeatedly traverses tables up to~\unit[1]{m} high, 
performs more than 15 consecutive autonomous jumps while following waypoints, 
handles previously unseen multi-table configurations, 
executes continuous repertoires of kip-ups, jumps, flips, kip-downs, 
over more than 20 consecutive trials, 
and performs more than 10 consecutive autonomous and steerable 
repertoires of wheelies, lateral jumps, and single-wheel jump downs.
These results demonstrate that RL can 
endow bicycle robots with levels of agility previously 
associated primarily with legged platforms while preserving the speed 
and efficiency of wheeled locomotion, 
establishing a foundation for bicycle acrobatics.
Project page: \href{https://shamelfahmi.com/bicycle_acrobatics/}{shamelfahmi.com/bicycle\_acrobatics}
\end{abstract}
\maketitle
\section*{Introduction}
Despite the reliability of legged robots in unstructured environments, 
their practical utility in long-range missions is limited by high energetic costs and low operational speeds. 
While quadrupeds and humanoids excel at traversing obstacles, 
maintaining high speeds typically incurs a severe penalty on battery life. 
This trade-off is 
challenging for applications 
where robots must traverse long distances quickly and efficiently. 
Traditional wheeled robots solve the efficiency problem 
but are fundamentally halted by curbs, stairs, or boxes. 
A persistent gap thus remains between speed, agility, and traversability.
To address this gap, we introduced the~\gls{umv}~\cite{RAIInstitute2026UMV}, 
a bicycle robot that pairs the efficiency of wheeled robots with the athleticism of legged ones 
by combining bicycle-like locomotion with dynamic articulation.

Our previous work introduced~\gls{umv} and presented its system design~\cite{RAIInstitute2026UMV}. 
Here, we focus on controlling bicycle robots like~\gls{umv} with~\gls{rl}, 
answering two questions:
how can we perform bicycle acrobatics with~\gls{rl}, 
and 
how can we orchestrate multiple acrobatic stunts into a long-repertoire demonstration?
We show that the designed~\gls{rl} behaviors enable 
various stunts including 
jumps, flips, wheelies, bunny hops, three-point turns, driving, and kips.

\subsection*{Bicycle Robots: Advantages, Challenges, and Related Work}
Compared to legged and hybrid wheeled-legged 
robots~\cite{boston2017handle, kashiri2019centauro, klemm2019ascento, bjelonic2020rolling, bjelonic2022survey, liu2024diablo, unitree2025bw}, 
which use multi-track configurations with wheels arranged side by side, 
bicycle robots like~\gls{umv}
have a single-track configuration with wheels in tandem.
This yields a simpler mechanical design with a smaller footprint and fewer~\gls{dofs}
that result in 
lower mass, 
higher energy efficiency and speed, 
reduced costs, 
and a lower risk of mechanical failure.

The unique and simple design of bicycle robots like~\gls{umv} 
comes at the cost of control.
Controlling a bicycle 
presents more challenges compared to a legged robot for several 
reasons~\cite{jones2006archives, astrom2005balance, meijaard2007linearized, kooijman2011, cui2020nonlinear, xiong2024steering, xiong2024steering2}. 
First, legged robots have multiple points of contact and a ``support polygon'': 
quadrupeds and humanoids can remain statically stable when their joints are locked~\cite{cui2020nonlinear}. 
Bicycle robots instead have a line of contact, 
making them statically unstable and more difficult to balance or 
track-stand (stand in place)~\cite{cui2020nonlinear, astrom2005balance, jones2006archives}. 
Second, bikes have non-holonomic driving constraints and non-minimum phase dynamics: 
turning right at high speed requires first steering momentarily left, 
a counter-intuitive effect known as counter-steering~\cite{xiong2024steering}. 
Third, bicycles have far fewer~\gls{dofs}. 
A humanoid can use ankle torque or swing its arms to arrest a fall, 
and a quadruped can step out to form a new stable base, 
whereas a bicycle robot cannot instantaneously change 
its base of support
to catch a fall but must navigate a curve to do so, 
making track-standing difficult. 

Bicycle control has long been a benchmark 
for both model-based and learning-based control. 
Model-based and optimization approaches have been investigated to balance and steer 
bicycles~\cite{xiong2024steering, cui2020nonlinear, wang2024bayesian, wang2024attitude}.
Early applications of~\gls{rl} showed that 
appropriately designed reward functions suffice to solve elementary riding behaviors~\cite{randlov1998learning}, 
and more recent efforts extended~\gls{rl} to increasingly challenging 
environments~\cite{zhu2023deep, zheng2023reinforcement, baltes2023deep, zheng2022continuous, yuan2024dynamic}. 
Pushing the limits of bicycle agility, 
Tan et al.~\cite{tan2014stunts} 
used an offline policy-search approach to learn acrobatic behaviors 
including bunny hops, wheelies, endos, etc.
Although this prior work demonstrated a wide range of stunts, 
most results were evaluated only in simplified simulation environments, 
without validation in more realistic physics simulation or on physical robots.
Our work tackles highly dynamic behaviors of substantially greater complexity, 
validates them extensively on~\gls{umv} through many repeated trials, 
and orchestrates them into coordinated motions.

\subsection*{Reinforcement Learning for Locomotion}
Model-based control was long the dominant paradigm for 
locomotion~\cite{raibert1986legged, kuindersma2016optimization, dicarlo2018dynamic, bellicoso2018dynamic, fahmi2019passive}. 
Over the past decade, however,~\gls{rl} has emerged as a more powerful alternative. 
Thanks to large-scale parallel training~\cite{nikita2021learning} and sim-to-real,~\gls{rl}
has become the standard method for locomotion control of 
legged, wheeled, and hybrid robots~\cite{ha2025learning, bjelonic2022survey}, 
consistently outperforming classical model-based methods 
in robustness, adaptability, and behavioral complexity.

Early foundational work showed that locomotion policies trained entirely 
in simulation could transfer to physical quadrupeds, 
producing robust controllers that adapt to disturbances and terrain 
variations~\cite{nikita2021learning, lee2020learning}. 
Subsequent research increased the agility and robustness of learned behaviors. 
Miki et al.~\cite{miki2022learning} introduced student-teacher training with privileged information 
for fast, reliable locomotion on ANYmal, while 
Hoeller et al.~\cite{hoeller2024anymal}, Cheng et al.~\cite{cheng2024extreme}, 
and Zhuang et al.~\cite{zhuang2023robot} extended these capabilities to parkour-style locomotion.
These 
works~\cite{nikita2021learning, lee2020learning, miki2022learning, zhuang2023robot, hoeller2024anymal, cheng2024extreme} 
and others~\cite{li2023robust, lee2024learning, kim2025high, he2025attention, bussola2025guided} 
demonstrate the progression from robust locomotion 
to highly dynamic, athletic behaviors
such as jumping across large gaps, climbing tables, and traversing narrow surfaces.
We refer to these methods as~\emph{vanilla}~\gls{rl} 
(or as reference-free or tabula rasa~\gls{rl})
where the goal is to 
control and track the SE2 twist of the robot's base 
or to follow a waypoint. 
\emph{Waypoint following} 
has been specifically crucial for parkour-style tasks.

The main challenge with~\emph{vanilla}~\gls{rl} is that
it may require extensive reward engineering and exploration to discover complex behaviors.
For instance, encoding a maneuver such as a flip directly through reward shaping is nontrivial and often impractical.
\emph{Motion imitation} addresses this 
by incorporating expert demonstrations as reference trajectories. 
The~\emph{DeepMimic} framework~\cite{peng2018deepmimic} showed that~\gls{rl} 
combined with rewards that track motion from reference trajectories can reproduce complex behaviors such as flips. 
Follow-up work extended this to multiple reference trajectories~\cite{peng2020learning, peng2021amp, peng2022ase}. 
Recent approaches including
MaskedMimic~\cite{tessler2024maskedmimic}, 
SoftMimic~\cite{margolis2025softmimic}, 
and other variants~\cite{kang2025learning} 
improve robustness by 
allowing flexible rather than exact correspondence with demonstrations.
\emph{Motion imitation} has also been extended to diverse data sources beyond traditional~\gls{mocap}~\cite{peng2018deepmimic, peng2020learning}: 
videos~\cite{peng2018sfv}, 
model-based controllers~\cite{miller2023reinforcement, kang2023rl, youm2023imitating, fuchioka2022opt, liu2024opt2skill}, 
motions retargeted across morphologies~\cite{yang2024omniretarget}, and 
large motion datasets~\cite{mahmood2019amass, harvey2020robust}.
Leveraging these developments, 
state-of-the-art work on humanoid robots has acquired dynamic whole-body skills 
by combining~\gls{rl}, motion priors, and large-scale 
datasets~\cite{he2024omnih2o, he2025asap, chen2025gmt, wu2026perceptive, sleiman2026zest}.

One main challenge with \emph{motion imitation} is that it
assumes access to high-quality demonstrations covering the complete task. 
In practice, however, demonstrations may be noisy, imperfect, or incomplete. 
Several works overcome this challenge~\cite{brown2019extrapolating, wu2019imitation},
in particular, \gls{imi}~\cite{kim2026flip}, which shows that~\emph{motion imitation}
can be effective despite imperfect references that can initially be kinematically or dynamically infeasible. 
\gls{imi} is especially important for bicycle robots and other unconventional morphologies, 
where obtaining high-quality reference motions is impractical; 
unlike humanoids or quadrupeds, 
these platforms lack large~\gls{mocap} or video collections, 
retargeting from other embodiments is difficult, 
and designing full kinematic reference trajectories is hard.

\emph{Guided tracking}~\gls{rl} is a trade-off between \emph{vanilla}~\gls{rl} and \emph{motion imitation}, 
which does not need full-state references and instead uses partial demonstrations
as a guide for the policy to track.
Several works~\cite{li2023learning,zargarbashi2024} 
showed that even incomplete trajectory segments provide useful guidance, 
substantially reducing the burden of collecting full-state demonstrations.
Particularly important to this work is LineRides~\cite{rho2026linerides}.
Rather than relying on full-state demonstrations, 
LineRides only needs a simple sketch 
or line cues that capture the essential structure of a maneuver
while remaining easy to design enabling challenging bicycle stunts on~\gls{umv}. 

A final alternative is \emph{pose reaching}~\gls{rl}, 
where the objective is restricted to achieving a designated final 
pose~\cite{ma2023learning, huang2025learning, strauch2025robot}. 
Rather than following a command or trajectory, 
the policy is conditioned on this terminal state, 
granting the agent flexibility to determine its own execution 
path
across varying robot configurations.

Taken together, these families of \gls{rl} methods can be ordered along a spectrum 
distinguished by reward density and the complexity of the reference or command. 
\emph{Motion imitation} occupies the dense end due to its reliance on state-level references, 
\emph{waypoint following} via \emph{vanilla}~\gls{rl} occupies the sparse end, 
while SE2 twist tracking, pose reaching, and guided tracking lie between these extremes.
In contrast to this prior work on quadrupeds and humanoids, 
we focus on bicycle robots in this work. 

\subsection*{Proposed Approach and Contributions}
Achieving highly dynamic stunt behaviors on bicycle robots presents two fundamental and unexplored challenges. 
First, while~\gls{rl} has enabled complex motor skills across legged robots, 
there has been no systematic effort to 
train~\gls{rl} policies that can be reliably deployed on bicycle robots, 
neither in physically realistic simulation nor on physical hardware. 
Hence, learning dynamic stunt-like maneuvers on underactuated, 
nonholonomic systems such as bicycles remains an open problem. 
Second, even as modern RL methods can produce diverse individual behaviors, 
composing them into a coherent, long-horizon stunt sequence remains a major challenge, 
and existing approaches typically demonstrate a single stunt rather than a long repertoire. 
Yet real-world bike acrobatics, 
such as those popularized in videos by Danny MacAskill~\cite{macaskill2013imaginate, macaskill2009streettrials}, 
shows that impressive feats arise not from isolated tricks 
but from the seamless orchestration of many behaviors in a single continuous run. 
Although such performances may be rehearsed offline over multiple takes, 
their compelling nature stems from
the robust and reliable execution under environmental variation.

Inspired by this paradigm, 
we develop a method for learning and composing highly dynamic acrobatic bicycle stunts using~\gls{rl} 
that addresses both challenges jointly. 
First, 
we design and train~\gls{rl} controllers that generate
agile bicycle behaviors in simulation and transfer them to hardware, 
exploring multiple~\gls{rl} formulations to enable stable learning on 
this inherently unstable platform 
and developing training procedures tailored to each stunt.
Second, 
we introduce 
an~\emph{orchestrator} that composes 
individually trained skills into extended, 
continuous stunt repertoires, inspired by~\cite{burridge1999sequential}, 
with mechanisms for sequencing, transitioning, 
and coordinating policies to enable long-horizon 
execution within a single uninterrupted run.
Third, 
we validate our algorithms on~\gls{umv}, 
demonstrating unprecedented agility in bicycle robotics and, 
long-horizon repertoires of composed acrobatic behaviors executed in a single run. 
Together, our contributions close a critical 
gap in~\gls{rl} control for underactuated wheeled systems 
and establish a foundation for expressive, high-performance autonomous bicycle robots.


\section*{Results}
\begin{table}[t]
\centering
\caption{List of Stunts Performed by UMV}
\label{tab_robot_stunts_taxonomy}
\includegraphics[width=0.95\columnwidth]{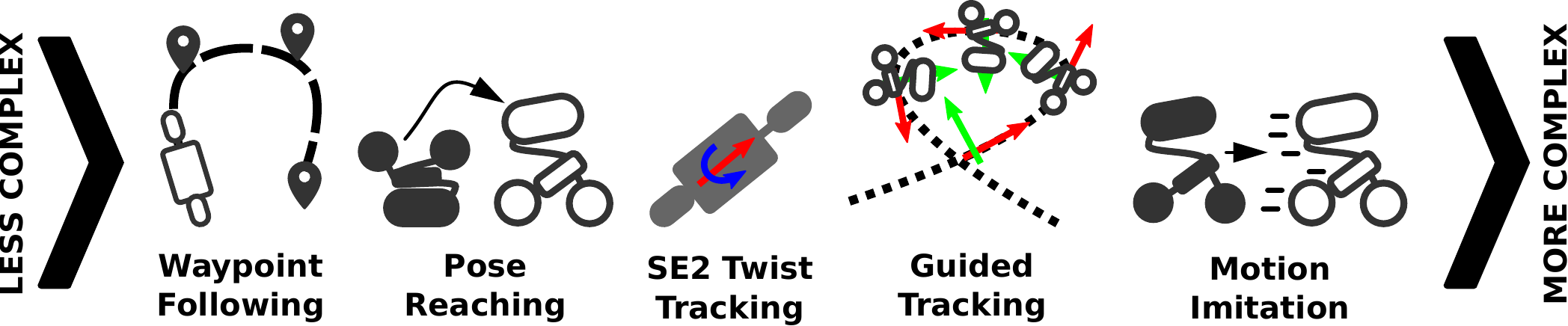}
\scriptsize{    
\renewcommand{\arraystretch}{1.4} 
\begin{tabular}{lcc}
\midrule
\textbf{Stunts} & \textbf{Policy $\pi$}      & \textbf{Success Rate} \\ \midrule
\ourrow \textbf{(A) Waypoint Following}      & \multicolumn{2}{r}{(command dimension: $2D$)}\\
1. Navigation                                & $\pi_j$                                           & -- \\
2. 75cm \& 1m Jump                           & $\pi_j$                                           & 0.94 \\  
3. Two Table Jump                            & $\pi_j$                                           & 0.67 \\  
4. Steerable Jumps                           & $\pi_j$                                           & 0.83 \\  
5. Three Table Jump                          & $\pi_{mj}$                                        & 0.50 \\  
6. Lateral Jumps (pre-training)$^1$          & $\pi_{lj0}$                                       & -- \\
7. Autonomous Wheelie$^1$                    & $\pi_{aw}$                                        & -- \\ 
\ourrow \textbf{(B) Pose Reaching} & \multicolumn{2}{r}{(reference dimension: $SE3$)}\\
8. Kip Up                                    & $\pi_{ku}$                                        & -- \\
9. Kip Down                                  & $\pi_{kd}$                                        & -- \\
\ourrow \textbf{(C) SE2 Twist Tracking} & \multicolumn{2}{r}{(command dimension: $SE2$)}\\
10. Driving                                  & $\pi_d$                                           & 1.00 \\ 
11. Steerable Wheelie$^1$                    & $\pi_{sw}$                                        & 1.00 \\
\ourrow \textbf{(D) Guided Tracking}         & \multicolumn{2}{r}{(reference dimension: $SE3 \times M$)}\\
12. Bunny Hops                               & $\pi_{h}$                                         & -- \\
13. Three-point Turns                        & $\pi_{tpt}$                                       & -- \\ 
\ourrow \textbf{(E) Motion Imitation}        & \multicolumn{2}{r}{(reference dimension: $N \times M$)}\\
14. Flips                                    & $\pi_f$                                           & -- \\
15. Lateral Jumps                            & $\pi_{lj}$                                        & -- \\
\ourrow \textbf{(F) Orchestrated Stunts}     & ~                                                 & ~ \\
16. Jump \& Track Stand                      & $\pi_j \land \pi_d$                               & -- \\ 
17. Flip \& Drive                            & $\pi_d \land \pi_f$                               & 1.00 \\
18. Three-point Turns \& Drive               & $\pi_d \land \pi_{tpt}$                           & 1.00 \\ 
19. Hop \& Drive                             & $\pi_d \land \pi_{h}$                             & 1.00 \\ 
20. Kip \& Drive                             & $\pi_{ku} \land \pi_d \land \pi_{kd}$             & 1.00 \\
21. Kip, Jump \& Flip                        & $\pi_{ku} \land \pi_j \land \pi_f \land \pi_{kd}$ & 0.97 \\ 
22. Autonomous Wheelie \& Lateral Jump$^1$   & $\pi_{aw} \land \pi_{lj} \land \pi_{j}$           & 0.93 \\ 
23. Steerable  Wheelie \& Lateral Jump$^1$   & $\pi_{sw} \land \pi_{lj} \land \pi_{sw}$          & 0.92 \\ 
24. Aut. Wheelie \& Two Table Lat. Jump$^1$  & $\pi_{aw} \land \pi_{lj} \land \pi_{j}$           & 0.67 \\ 
25. Ste.  Wheelie \& Two Table Lat. Jump$^1$ & $\pi_{sw} \land \pi_{lj} \land \pi_{sw}$          & 0.67 \\ 
26. Park From Anywhere                       & $\pi_j \lor \pi_d$                                & --\\ 
\midrule
\multicolumn{3}{l}{$^1$ These stunts are using the~\gls{umv} version with the out-of-sagittal-plane Yoll joint.}
\end{tabular}
}
\end{table}

\begin{figure}
\centering
\includegraphics[width=\columnwidth]{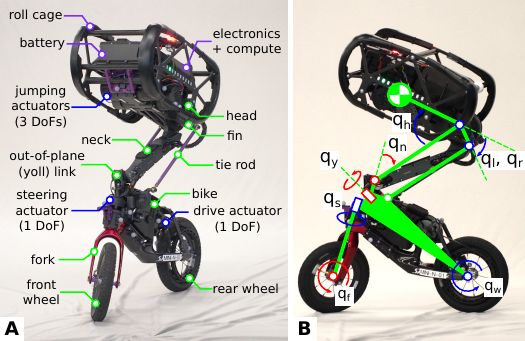}
\caption{\textbf{Ultra Mobility Vehicle (UMV).}
\textbf{(A)} 
The robot has eleven rigid links (green), five actuators (blue), and additional structural components (purple). 
The eleven links are the \emph{Head}, left and right \emph{Fins}, \emph{Neck}, 
left and right \emph{Tie Rods}, \emph{Bike}, \emph{Fork}, 
out-of-plane \emph{Yoll}, and front and rear \emph{Wheels}. 
\textbf{(B)}
The five actuated joints are shown in blue, and the passive joints are shown in red. 
The joints angles are the
\emph{Head} $q_h$, 
\emph{Left Fin} $q_l$, 
\emph{Right Fin} $q_r$, 
\emph{Neck} $q_n$, 
\emph{Yoll} $q_y$, 
\emph{Steering} $q_s$, 
\emph{Front Wheel} $q_f$, and 
\emph{Rear Wheel} $q_w$.
}
\label{fig_umv_overview}
\end{figure}

\subsection*{Overview}
\begin{figure*}
\centering
\includegraphics[width=2\columnwidth]{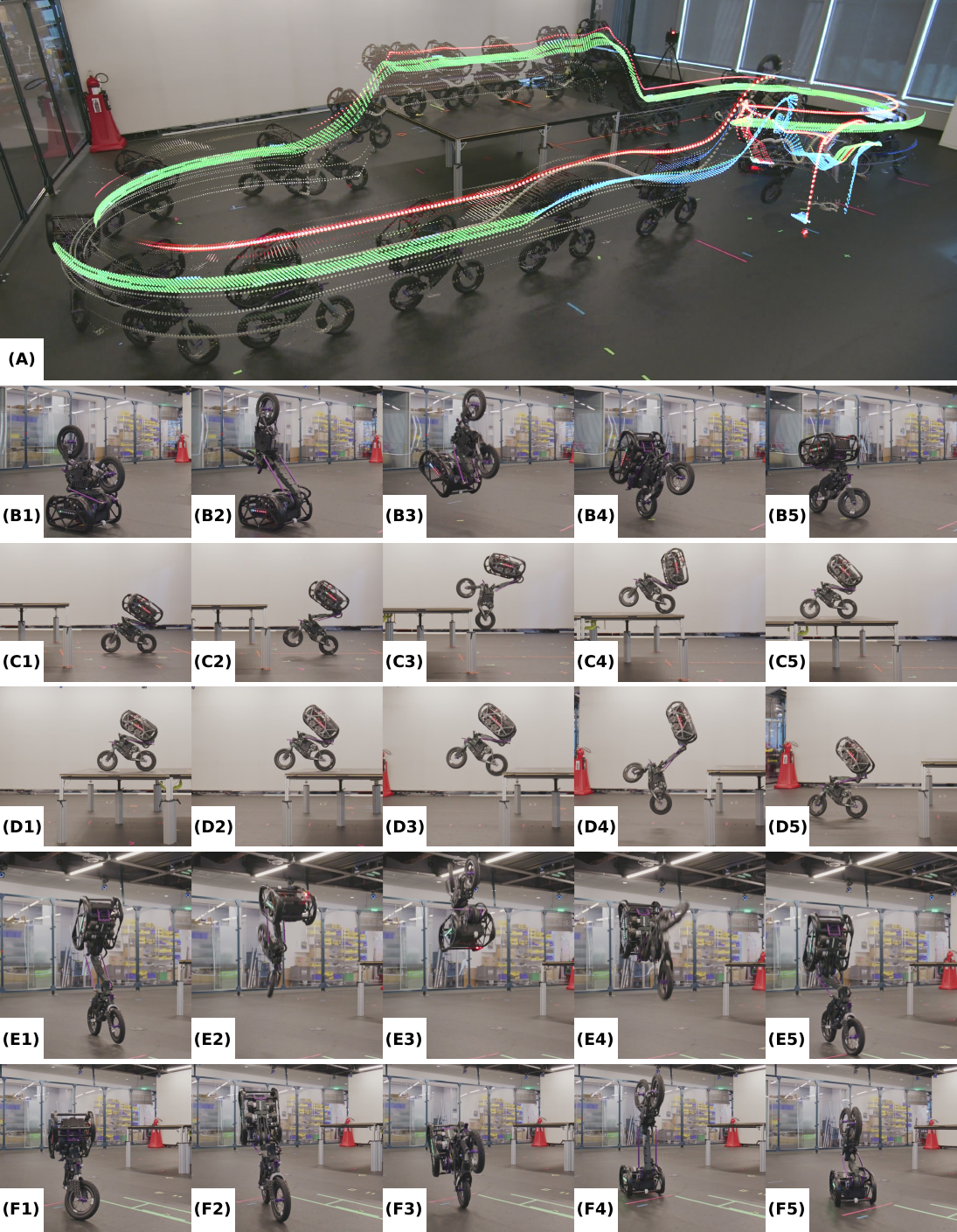}
\caption{\textbf{Orchestrated Kips, Jumps, \& Flips.} 
The robot's trace (\textbf{A}) and corresponding key video frames (\textbf{B--F}) 
illustrate a single long orchestrated sequence of stunts on~\gls{umv}. 
The stunts are performed seamlessly in the following order: 
(\textbf{B}) kip-up, 
(\textbf{C}) table jump up 
(\textbf{D}) table jump down,
(\textbf{E}) front flip,
(\textbf{F}) kip-down.
This repertoire is robustly demonstrated across more than 20 consecutive trials.
}
\label{fig_collage_odnt}
\end{figure*}

\begin{figure*}
\centering
\includegraphics[width=2\columnwidth]{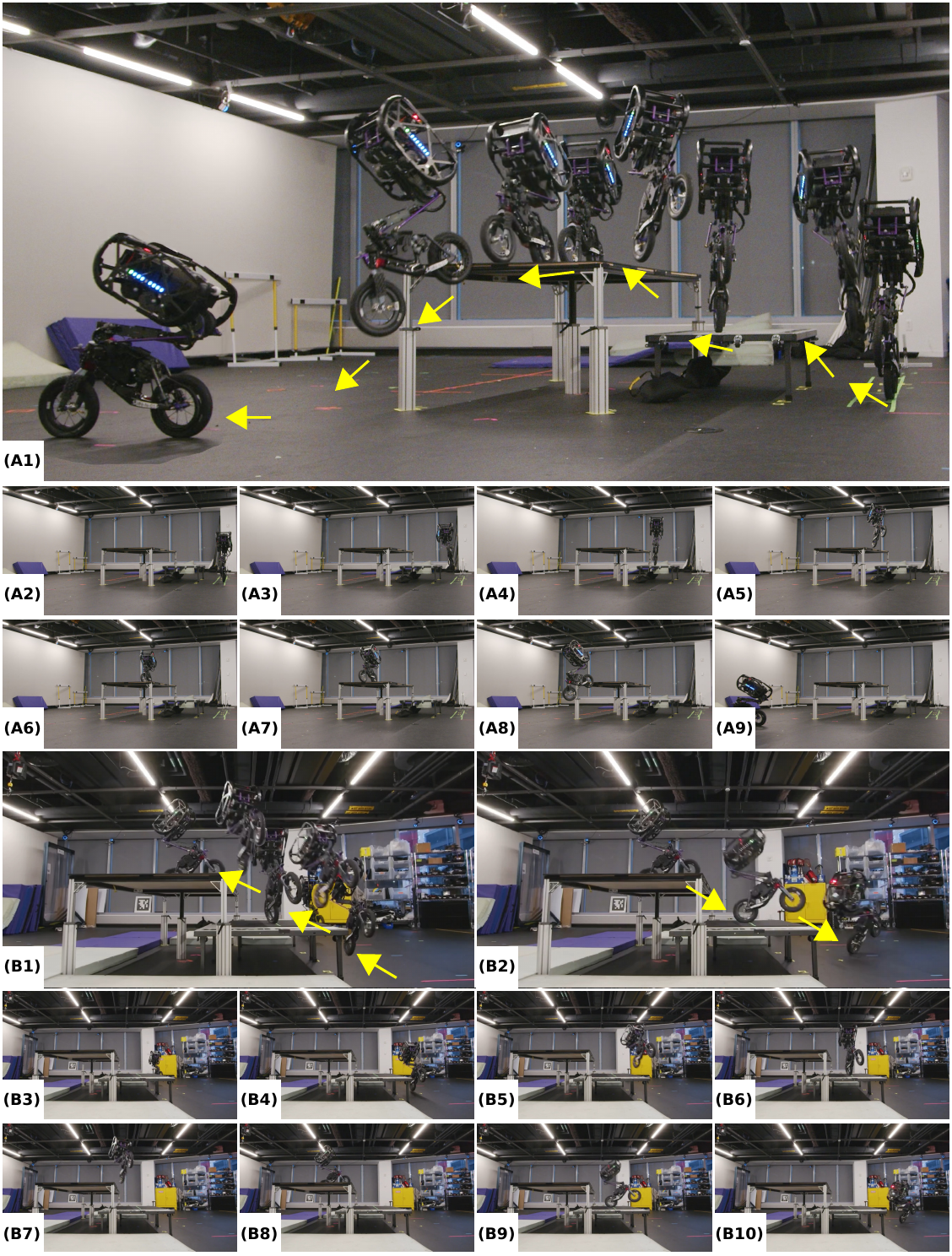}
\caption{\textbf{Orchestrated Lateral Jumps \& Wheelies.} 
The robot's trace and corresponding key video frames of autonomous (\textbf{A}) and 
steerable (\textbf{B}) wheelie and lateral jump stunts.
The robot performs (\textbf{B}) solely by balancing on the rear wheel. 
Yellow arrows indicate the direction of motion.
}
\label{fig_demo2}
\end{figure*}

\begin{figure}
\centering
\includegraphics[width=0.99\columnwidth]{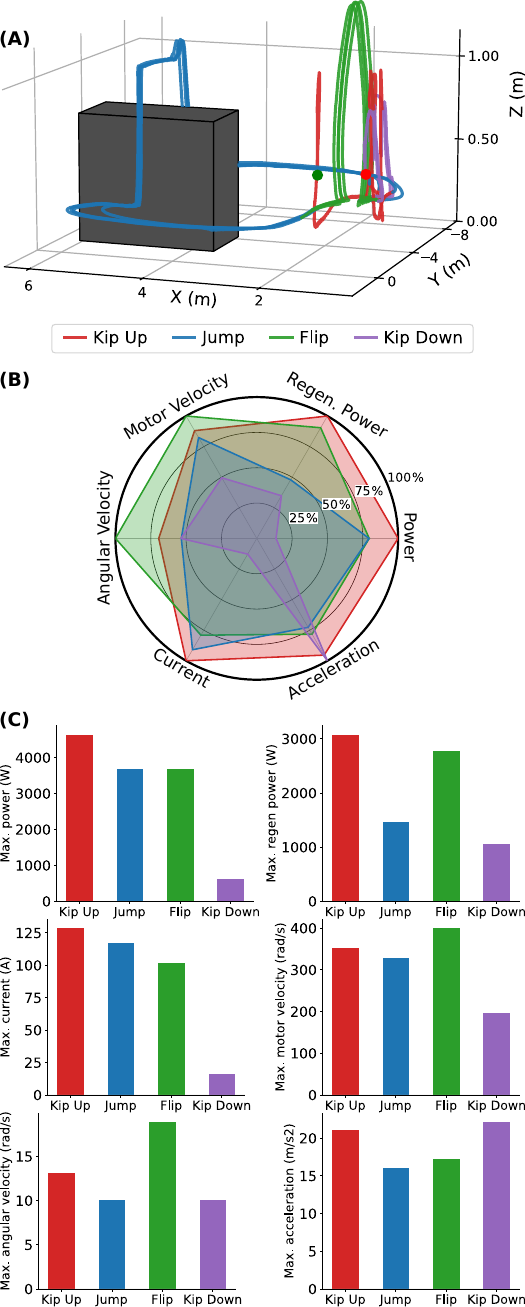}
\caption{\textbf{Performance Analysis of the Kips, Jumps \& Flips.}
(\textbf{A})~3D trajectory of four consecutive loops. 
(\textbf{B})~Radar chart of key metrics across the four stunts. 
(\textbf{C}--\textbf{H})~Bar charts of the absolute peak values of these metrics across the stunts.
The metrics are maximum power, regeneration power,  motor velocity, angular velocity, current, and linear acceleration. 
}
\label{fig_odnt}
\end{figure}

\begin{figure*}
\centering
\includegraphics[width=2\columnwidth]{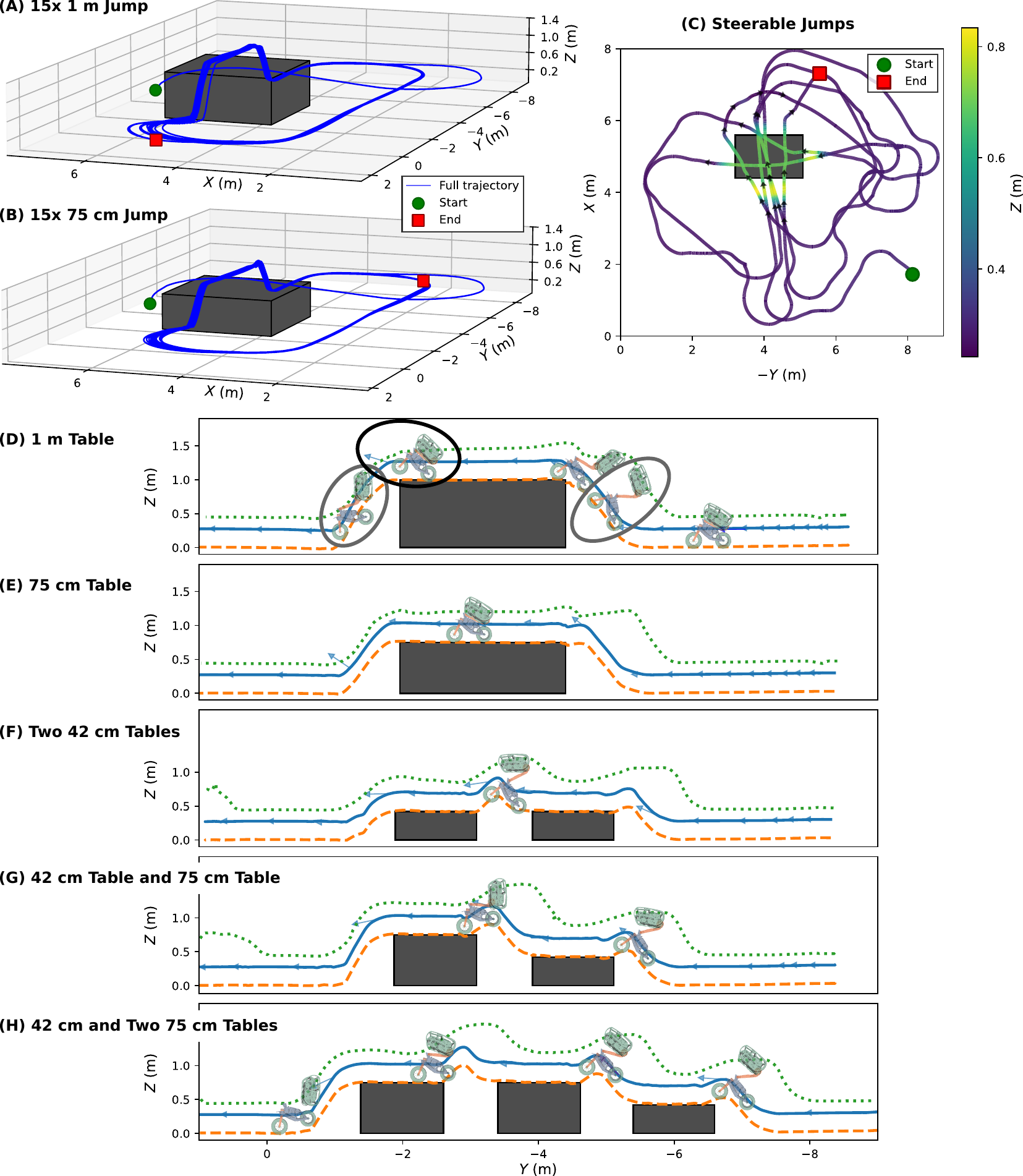}
\caption{\textbf{Autonomous \& Steerable \& Table Jumps Analysis.}
(\textbf{A})~15 times autonomous \unit[1]{m} table jump with a single policy. 
Green circles and red squares indicate start and end positions. 
Solid lines denote full trajectories.
(\textbf{B})~15 times \unit[75]{cm} table jump with the same policy.
(\textbf{C})~Top view of a steerable jump while tracking continuously varying waypoints. 
Trajectory color indicates robot height $Z$.
(\textbf{D})~Side view of a \unit[1]{m} table jump. 
Solid, dashed, and dotted lines denote body, clearance, and upper-body trajectories. 
Gray ellipses highlight emergent snake-like climbing during takeoff and landing. 
Black ellipse highlights a wheelie-like posture before descent.
(\textbf{E})~Side view of a \unit[75]{cm} table jump.
(\textbf{F})~Side view of two \unit[42]{cm} tables jump using a policy trained only on single-table environments.
(\textbf{G})~Side view of a \unit[42]{cm} table jump followed by a \unit[75]{cm} table.
(\textbf{H})~Side view of a \unit[42]{cm} table jump followed by two \unit[75]{cm} tables jumps.
The robot moves from right to left in all side-view panels.
}
\label{fig_all_jumps}
\end{figure*}

\begin{figure*}
\centering
\includegraphics[width=2\columnwidth]{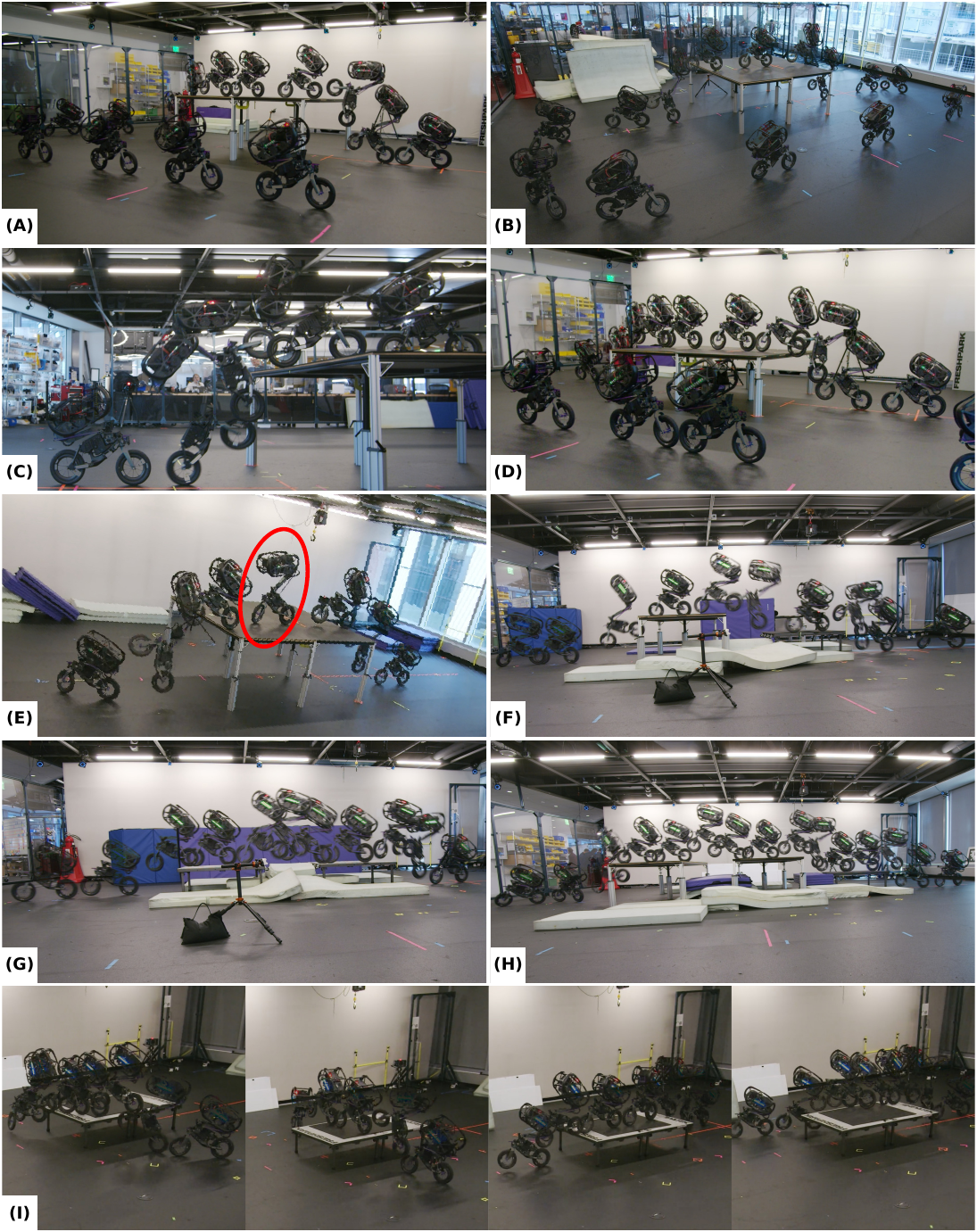}
\caption{\textbf{Autonomous \& Steerable \& Table Jumps Stunts.} 
The figures present a collage of snapshots from the various experiments 
demonstrating the diverse set of jumping stunts and long repertoires achieved by~\gls{umv}.	
In order of appearance from top left to bottom right, the figure illustrates the following behaviors. 
(\textbf{A,B}) 1 m jumps from two different views, 
(\textbf{C,D}) 75 cm jumps from two different views, 
(\textbf{E}) a jump and track stand on top of the table.
(\textbf{F,G}) two-table jumps with different heights, 
(\textbf{H}) a three table jump with different heights, and
(\textbf{I}) joystick-steerable jumps over a table.
}
\label{fig_collage_jumps}
\end{figure*}

We demonstrate a diverse set of acrobatic and dynamic maneuvers performed by~\gls{umv}, 
learned using several~\gls{rl} formulations 
ranging from~\emph{waypoint following} to \emph{motion imitation}.
All stunts are summarized in~\tref{tab_robot_stunts_taxonomy} 
and described in detail in the Methods section. 
\rev{Movie~S1} overviews the approach and showcases the experiments discussed below. 
All results are obtained on the physical robot. 
The stunts in~\tref{tab_robot_stunts_taxonomy}(A--E) 
are executed using single policies, 
whereas those in~\tref{tab_robot_stunts_taxonomy}(F) require orchestrating multiple policies; 
the policies used in each experiment are specified in the corresponding subsection. 
Figures~\ref{fig_collage_odnt}--\ref{fig_other_stunts} present representative snapshots 
illustrating the range of stunts achieved.
To find the exact breaking points of our policies and controllers, 
we ran multiple ablation studies, 
detailed in~\sref{sec_ablations}. 

The \gls{umv} robot shown in~\fref{fig_umv_overview} 
has eleven links: 
Head, 
left and right Fins, 
Neck, 
left and right Tie Rods, 
Bike, 
Fork, 
out-of-plane Yoll, 
and front and rear Wheels, 
and seven main joints:
the~\emph{Head}~($q_h$), 
\emph{Left Fin}~($q_l$), 
\emph{Right Fin}~($q_r$), 
\emph{Neck}~($q_n$), 
\emph{Yoll}~($q_y$), 
\emph{Steering}~($q_s$), 
and \emph{Rear Wheel}~($q_w$). 
The joints 
$q_h$, 
$q_l$, 
$q_r$, 
$q_s$, 
and~$q_w$ are actuated while 
the remaining ones are passive.
Forward motion comes from the rear wheel drive~$q_w$, and steering from~$q_s$. 
The remaining three~\gls{dofs}, $q_h$, $q_l$, and $q_r$, 
control the position and orientation of the \emph{Head} 
relative to the \emph{Bike} through a spatial linkage 
consisting of the \emph{Neck} link and left and right \emph{Tie Rods}. 
Differential motion between $q_l$ and $q_r$ produces out-of-sagittal-plane movement 
that we refer to as the \emph{Yoll} motion
since the joint partly yaws and partly rolls. 
Some versions of~\gls{umv} have the \emph{Yoll} joint \emph{physically locked}
where $q_l = q_r$ (yielding $q_y = 0$).
We refer the reader to~\sref{sec_umv_sup} for further details.

\subsection*{Orchestrated Acrobatic Stunts}
The \emph{Kip, Jump \& Flip} stunt in~\tref{tab_robot_stunts_taxonomy}(21) and \fref{fig_collage_odnt}
is a long-repertoire acrobatic stunt where the robot starts in an upside-down crouched configuration. 
A user triggers a button for the robot to kip up and follow several waypoints. 
The robot loops through the sequence and, once it finds a table, jumps onto and off it.
On the way back it performs a flip, continues following the waypoints, and kips down at the end. 
This stunt uses four policies, \textit{kip up}~$\pi_{ku}$, \textit{jump}~$\pi_j$, \textit{flip}~$\pi_f$, 
and \textit{kip down}~$\pi_{kd}$,  
and the orchestrator dynamically transitions between them. 
The orchestrator decides when to switch from one policy to another based on the state of the robot, 
its location, and commands coming from a user to specify when to kip up or kip down. 
This state-based switching enables the robot to recover from real-world variations between stunts, 
ensuring that the termination state of one policy conditions the initialization of the next.

Figure~\ref{fig_collage_odnt} shows screenshots of this experiment. 
Figure~\ref{fig_collage_odnt}(A) shows a trace of one run in which the robot 
loops through the sequence five times, 
and Figures~\ref{fig_collage_odnt}(B--E) show a single trial of the stunts. 
\rev{Movie~S2} shows more than 20 consecutive successful trials. 
Figure~\ref{fig_odnt} analyzes the performance of~\gls{umv} during a trial of five consecutive runs 
to determine which of the four stunts is most demanding. 
We show the robot's trace, color-coded by stunt, in~\fref{fig_odnt}(A), 
and evaluate the following metrics across the five trials and four stunts:
maximum total mechanical joint power, maximum total mechanical joint regeneration power, 
maximum total actuator motor velocity, maximum total bus current, maximum total bike angular velocity, 
and total bike linear acceleration. 
These normalized metrics are compared qualitatively in the radar chart of~\fref{fig_odnt}(B) 
and quantitatively in the bar charts of~\fref{fig_odnt}(C--H).
Figure~\ref{fig_odnt} 
reveals that the \textit{kip up} demands the highest 
overall peak power ($\approx \unit[4600]{W}$) 
and bus current ($\approx \unit[130]{A}$) of any maneuver. 
This aligns with the mechanics of the stunt: 
unlike others that build on forward driving momentum, 
the \textit{kip up} must launch the robot from a stationary, upside-down position. 
As a result, all of the momentum required to establish a standing stance 
must be generated through immediate explosive torques from the body motors.
In contrast, 
the \textit{flip} leverages kinetic energy gained during the preceding drive phase, 
so its power and current demands are lower, 
yet it registers the maximum angular velocity ($\approx \unit[19]{rad/s}$) 
because the robot must build up angular momentum fast to turn 360 degrees in a short time. 
The \textit{jump} exhibits a balanced, high-demand profile, 
with significant power output during launch and a high 
regeneration power upon landing. 
This high regeneration power indicates that the actuators actively back-drive 
to absorb and damp the impact, as expected when jumping from a high table.
The \textit{kip down} represents the lowest electrical load on the system, 
registering minimal power, current, and motor velocity. 
This is expected since the robot essentially
is falling backwards but in a controlled way. 
Because it is hard to control falling with~\gls{umv}'s few~\gls{dofs}, 
the \textit{kip down} logs the highest peak linear acceleration magnitude~($\approx \unit[22]{m/s^2}$). 

The~\emph{Wheelie \& Lateral Jump} stunts in~\tref{tab_robot_stunts_taxonomy}(22--25)
are another example of a long-repertoire acrobatic stunt where the robot first wheelies towards a table, 
jumps sideways on top of one table, balances on one wheel, jumps sideways on top of another table, 
and then wheelies down. 
This stunt requires the full system with the unlocked \emph{Yoll} joint. 
\fref{fig_demo2}~shows two versions of this acrobatic stunt
where the robot was either autonomous~(\fref{fig_demo2}(A)) or user-driven~(\fref{fig_demo2}(B)).
In the fully autonomous case, 
an autonomous wheelie policy~$\pi_{aw}$ drives the robot to a waypoint and aligns it with the table.
An imitated lateral jump policy~$\pi_{lj}$ is then triggered, 
followed by a wheelie realignment~$\pi_{aw}$, a second imitated lateral jump~$\pi_{lj}$, 
and finally a waypoint-following jump-down policy~$\pi_j$. 
In the steerable case, 
a user manually steers a blind single-wheel wheelie policy~$\pi_{sw}$ to align the robot, 
triggers an imitated lateral jump~$\pi_{lj}$, realigns with~$\pi_{sw}$, performs a second imitated lateral jump~$\pi_{lj}$, 
and drives down using the same steerable wheelie policy~$\pi_{sw}$. 
\rev{Movie~S3} shows different trials of this stunt on including a one and two table jumps.

\subsection*{Table Jumps}
Figures~\ref{fig_all_jumps}(A,B) and~\ref{fig_collage_jumps}(A--D) show~\gls{umv} executing 
15 repeated jumps 
over single tables of \unit[75]{cm} and \unit[1]{m} heights
using the same single-table policy~$\pi_j$. 
In both experiments, 
the robot autonomously follows a sequence of user-defined waypoints 
without joystick control or manual intervention. 
The robot uses~\gls{mocap} to localize these waypoints and the table. 
As the robot follows the waypoints, it samples a robot-centric heightmap from a predefined one, 
and adapts its motion accordingly. 
The policy is not modified between the \unit[75]{cm} and \unit[1]{m} experiments, 
demonstrating successful deployment across different table heights, 
and highlighting the robustness of the jumping behavior 
as well as the robot's hardware, state estimation, and control pipeline. 

\fref{fig_all_jumps}(D,E) reveal several emergent behaviors. 
First, during both takeoff and landing, the robot exhibits a ``snake-like'' climbing motion
(gray ellipses in~\fref{fig_all_jumps}(D)). 
Rather than initiating a large ballistic jump far from the table, 
the robot approaches closely, pitches upward immediately before takeoff, 
and pitches downward just before touchdown. 
This appears to reduce unnecessary vertical thrust and smooth the transition onto and off the table. 
Since the policy penalizes excessive torque and power as well as hard impacts, 
this strategy emerges as an energy-efficient and mechanically favorable solution.
Second, near the end of the table traversal 
(black ellipse in~\fref{fig_all_jumps}(D)), 
the robot briefly performs a slight wheelie before descending. 
This behavior consistently emerged when training with noise and drift in the perceived heightmap, 
suggesting the policy learned a conservative strategy that prepares for small discrepancies 
between the estimated terrain and the actual table geometry, 
improving robustness to sensing or localization errors.

Figures~\ref{fig_all_jumps}(C) and \ref{fig_collage_jumps}(I)
show a steerable jumping behavior using the same jumping policy~$\pi_j$. 
The top view plots the robot's XZ position, with curve color indicating robot height 
and arrows indicating direction. 
Here, a user steers the robot by changing the waypoint dynamically via a joystick, 
so the robot follows a continuously changing waypoint. 
Unlike the previous experiments, 
where the robot only approaches the center of the table, 
in this experiment, 
the robot repeatedly approaches the table from different angles, performs the jump, 
and continues along nontrivial curved paths after landing. 
Combining agile jumping with steering substantially increases practical applicability 
and demonstrates that the policy captures a general locomotion strategy rather 
than a specialized stunt, without training separate policies for each.

Figures~\ref{fig_all_jumps}(F--H) 
and~\ref{fig_collage_jumps}(F--H)
show~\gls{umv} traversing multiple tables.
The same single-table policy~$\pi_j$ is used for both of the two-table jumps:
the two \unit[42]{cm} tables in~Figures~\ref{fig_all_jumps}(F)~and~\ref{fig_collage_jumps}(F) 
and the \unit[42]{cm} and \unit[75]{cm} tables in~Figures~\ref{fig_all_jumps}(G)~and~\ref{fig_collage_jumps}(G). 
The three-table jump in~Figures~\ref{fig_all_jumps}(H)~and~\ref{fig_collage_jumps}(H) 
proved too challenging for the single-table policy, 
so we trained a dedicated multi-table policy~$\pi_{mj}$ in a two-table setup. 
Deploying~$\pi_{mj}$ on three consecutive tables (\unit[42]{cm} and two \unit[75]{cm}) 
is therefore something that was not observed by the policy during training. 
In all of these experiments,
the policies are thus deployed on one more table than they were trained on. 
Yet,~\gls{umv} adapts online and completes the traversal without additional task-specific tuning.
Across all multi-table experiments, 
the side-view trajectories highlight the reliability of the~\gls{rl} policies. 
\gls{umv} consistently modulates its pitch and body posture while approaching and traversing tables, 
showing that the policies learn reusable dynamic strategies rather than memorizing a single jumping motion.
All the aforementioned experiments~(\tref{tab_robot_stunts_taxonomy}(1--5)) are shown in~\rev{Movie~S4}. 

\begin{figure*}
\centering
\includegraphics[width=1.8\columnwidth]{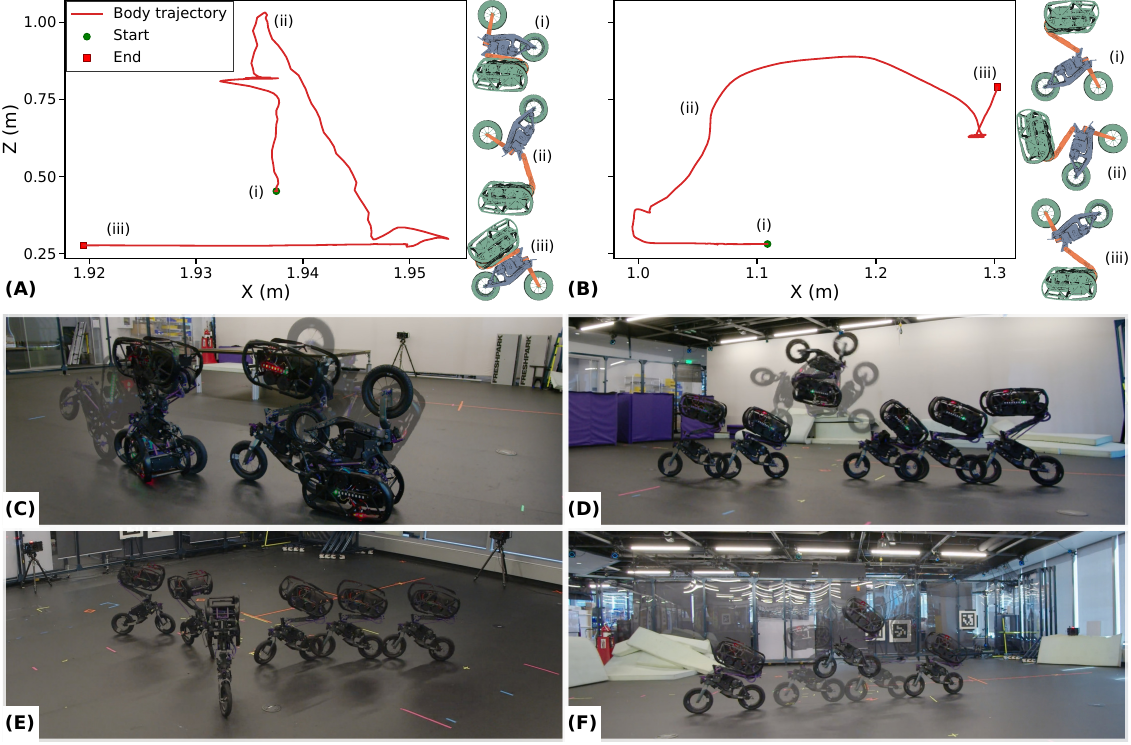}
\caption{\textbf{Additional Orchestrated Stunts.}
(\textbf{A,B})~Snapshots (i--iii) and the corresponding $X$--$Z$ body trajectory 
of the kip up~(\textbf{A}) and kip down~(\textbf{B}).
Overlaid snapshots of the
(\textbf{C})~kip up and down,
(\textbf{D})~front flip, 
(\textbf{E})~Three-point turn, 
and (\textbf{F})~bunny hop.
}
\label{fig_other_stunts}
\end{figure*}

\subsection*{Kip and Drive}
The \emph{kip and drive} stunt in~\tref{tab_robot_stunts_taxonomy}(20)
merges the~\emph{kip up}~$\pi_{ku}$, \emph{driving}~$\pi_d$, and~\emph{kip down}~$\pi_{kd}$ stunts 
in~\tref{tab_robot_stunts_taxonomy}(8, 10, and 9). 
As in the \emph{Kip, Jump \& Flip} stunt, 
the robot starts in an upside-down crouched configuration. 
Using a joystick, a user triggers a kip up, the robot transitions to driving, 
and another button triggers a kip down; this can repeat. 
Figures~\ref{fig_other_stunts}(A, B) show side views of the robot's bike position recorded from~\gls{mocap}
during kip up and kip down, respectively.
\fref{fig_other_stunts}(C) shows overlaid screenshots of a single trial, 
and~\rev{Movie~S5} show multiple trials. 

In the \textit{kip up} sequence (\fref{fig_other_stunts}(A)), 
the robot transitions from lying flat on its back to standing on its wheels: 
from a stationary resting position 
(i), it thrusts its bike base and wheels upward to a peak height just over \unit[1]{m} 
(ii), retracts the bike base to gain momentum, and 
then swings its wheels underneath to catch its weight, 
absorb the landing impact, and settle into a stable driving stance (iii). 
In the \textit{kip down} sequence (\fref{fig_other_stunts}(B)), 
the maneuver begins from the normal wheeled stance (i).  
Then, the robot falls backwards but in a controller way 
where it pitches the head up, almost vertical, over the rear wheel, before initiating a small wheelie 
to cause itself to fall backwards~(ii). 
Then, the robot lands it on its head frame, where it comes to rest (iii). 

\subsection*{Additional Stunts}
A key advantage of the orchestrator is its modularity. 
Rather than enforcing a fixed sequence of actions, 
it can seamlessly chain multiple behaviors, and it can enable on-the-fly operator intervention.
For example, 
an operator can interrupt the jumping policy to perform a track stand, 
as shown in~\fref{fig_collage_jumps}(E), 
before resuming autonomous execution.
We also use the orchestrator to merge standard driving 
with stunts from \emph{motion imitation}~\cite{kim2026flip} 
or \emph{guided tracking}~\cite{rho2026linerides}
as shown in~\rev{Movie~S6} and~\rev{Movie~S7}. 
A user drives the robot with normal steering inputs, 
and when a hot key is pressed, the orchestrator interrupts the drive policy, 
executes a targeted acrobatic maneuver, 
and seamlessly hands control back to driving once the stunt finishes. 
This enables stunts such as
\emph{front flips}~$\pi_f$,
\emph{three-point turns}~$\pi_{tpt}$, 
and \emph{bunny hops}~$\pi_{h}$, 
listed in~\tref{tab_robot_stunts_taxonomy}(17--19) and
shown in~\fref{fig_other_stunts}(D--F).

Parking a bicycle robot is a crucial stunt needed for robot testing. 
The \emph{kip down} stunt is one way to rest the robot at the end of an experiment, 
but sometimes we want the robot to navigate to a parking location first. 
To reduce this burden and the risk of end-of-run crashes, 
we developed an obstacle-aware navigation system for a \emph{park-from-anywhere} feature. 
With a single button press, 
the robot autonomously plans and executes a parking maneuver into the foam pile from its current location 
by searching an occupancy grid for a collision-free path. 
This high-level navigation stack provides 
either waypoints to the jump policy or twist commands to the drive policy. 
In \rev{Movie~S8}, we show 15 consecutive successful parking attempts.

\section*{Discussion}
Our results demonstrate that the highly dynamic, complex stunts
listed in~\tref{tab_robot_stunts_taxonomy} can be reliably executed on bicycle robots using~\gls{rl}. 
These behaviors are not one-off demonstrations as shown in the accompanying movies.
We ran these stunts several times on several robots without failures.
The demonstrated behaviors ranged from forward, lateral, autonomous, and steerable~\emph{jumps} 
to~\emph{kips}, \emph{flips}, \emph{hops}, and \emph{wheelies}. 
While bicycle robots inherently suffer from non-minimum phase dynamics and static 
instability compared to quadrupedal or humanoid counterparts, 
our framework shows that these control challenges can be overcome 
to unlock extreme agility.

A central finding is the effectiveness of modular policy orchestration. 
Rather than training a single monolithic policy, 
we categorized stunts by objective complexity and trained a specialist policy for each task. 
By tying transitions between policies, 
we enabled the robot to seamlessly chain 
behaviors~(e.g., \textit{Kip Up} to \textit{Drive}, 
\textit{Drive} to \textit{Jump}, and \textit{Flip} to \textit{Kip Down}) 
over continuous real-world deployments. 
The single-table jump, 
single lateral-table jump, 
and kip, jump \& flip stunts 
were successfully executed for over 15, 12, and 20 consecutive times, in a single run, respectively.
Our policies consistently recovered from 
the large contact impact with the ground inherent to these stunts 
and pushed the robot to its physical limits, 
including 
\unit[1]{m} table jumps, 
peak 
mechanical power of $\approx\unit[4600]{W}$ 
and bus currents of $\approx\unit[130]{A}$ during the \textit{kip up}, 
angular velocities of $\approx\unit[19]{rad/s}$ during the \textit{flip}, 
and linear acceleration of $\approx\unit[22]{m/s^2}$ during the \emph{kip down}.
The learned controllers were also resilient to terrain variation. 
The jumping policies adapted to table heights up to~\unit[1]{m} 
and compensated for spatial discrepancies without manually scripted corrections. 
The emergence of techniques such as the ``snake-like'' 
trajectory to minimize vertical thrust and wheelie-like postures for safe descent 
highlights the capacity of~\gls{rl} to 
discover energy-efficient, mechanically favorable solutions that preserve the hardware.

Despite these successes, several limitations remain. 
The current orchestration framework, while flexible, still
relies on explicit state boundaries and heuristic transitions to switch between controllers. 
A promising direction for future work is architectures
that can autonomously and continuously interpolate between agile behaviors without predefining these conditions.
Additionally, while \gls{umv} can robustly execute dynamic jumps and obstacle traversals, 
tasks that demand accurate wheel placement remain difficult. 
For instance,~\gls{umv} had low success rates executing any stunt with more than one table such as
the two-table and three-table forward jumps and two-table lateral jumps in~\tref{tab_robot_stunts_taxonomy}(3,5,24,25).
Addressing these tasks will likely require separating motion generation from motion execution. 
Instead of directly learning task-specific behaviors, 
a high-level motion generator would reason about the environment to synthesize and continuously update 
feasible reference trajectories online, while a general low-level tracking policy,
such as the guideline-tracking policy developed in this work, 
would execute these trajectories robustly. 
This hierarchy would allow online adaptation without sacrificing the robustness of the learned tracking controller.


\section*{Methods}
\subsection*{Training Details}
The bicycle stunts in this work~(\tref{tab_robot_stunts_taxonomy}) fall into 
five main categories 
based on the \emph{complexity} of their commands and references: 
\emph{Waypoint Following}, 
\emph{Pose Reaching}, 
\emph{SE2 Twist Tracking}, 
\emph{Guided Tracking}, and
\emph{Motion Imitation}.

The policies in~\tref{tab_robot_stunts_taxonomy}(A) 
such as the \emph{Jump}~$\pi_j$
and the \emph{Autonomous Wheelie}~$\pi_{aw}$ policies
are trained using \emph{Waypoint Following} 
where the objective is to reach a single target coordinate $x_\text{goal} \in \Rnum^{2}$. 
During training, the robot starts inside a ring (\fref{fig_ablations}(\textbf{A})) and the goal lies outside it. 
The policy observes a robo-centric heightmap, 
so jumping is emergent rather than prescribed.
Starting inside a ring, the robot discovers that the only way to reach the goal is to jump. 
Training without rings (using boxes) or with a single fixed ring height fails to converge. 
To solve this, we introduce a terrain-based curriculum based on~\cite{nikita2022advanced} 
and inspired by~\cite{xie2020allsteps, portelas2020automatic}, 
in which the robot begins on flat ground 
and the terrain difficulty (ring height) increases each time it succeeds in reaching the goal. 
During training, there is a single fixed waypoint, and during deployment, the robot accepts a list of waypoints. 
Since~$\pi_j$ is also trained on flat ground, 
we use it for driving to a waypoint as well as for jumping. 
We use~$\pi_j$ for the single- and two-table jumps, 
but it performs poorly in executing 
the three-table jump. 
We therefore train a multi-table jump policy~$\pi_{mj}$ 
with an identical setting except that the ring is replaced 
by two rings with a gap width that varies with terrain difficulty. 
More information is in~\sref{sec_training_jump}. 

The \emph{Kip Up}~$\pi_{ku}$ and \emph{Kip Down}~$\pi_{kd}$ policies 
in ~\tref{tab_robot_stunts_taxonomy}(B)
are trained with \emph{Pose Reaching}.
The objective is to reach a final robot pose $SE3\in \Rnum^{6}$ rather than a single point. 
The goal remains fixed during an episode.
This is more complex than waypoint following 
because the command is of higher dimensionality. 
The kip up policy is initialized in a fully tucked, upside-down configuration with the goal to stand upright, 
while the kip down policy is initialized upright with the goal to lay (kip) upside down. 
Both tasks use the full collision model, since the robot
utilizes the contact between the head and the ground.
These tasks are inspired by~\cite{huang2025learning, ma2023learning, hwangbo2019learning}. 
More information is in~\sref{sec_training_pose}. 

The~\emph{Driving}~$\pi_d$ and \emph{Steerable Wheelie}~$\pi_{sw}$ policies 
in~\tref{tab_robot_stunts_taxonomy}(C) use \emph{SE2 Twist Tracking}
where the objective is to track an $SE2$ twist command
of the bike's forward velocity~$v_c^x$, yaw rate~$\omega_c^z$, and zero lateral velocity. 
This is more complex than waypoint following and pose reaching
because the twist is a continuous command and can come from a joystick. 
During training the driving~$\pi_d$ policy, 
the SE2 twist command is
resampled after a certain time from a wide range of forward velocity and yaw rate. 

The \emph{Bunny Hops}~$\pi_{h}$ and 
\emph{Three-point Turns}~$\pi_{tpt}$ policies 
in~\tref{tab_robot_stunts_taxonomy}(D) 
are trained with \emph{Guided Tracking} where 
the objective is to track a series of~$M$ poses~$\in\Rnum^{SE3\times M}$
that guide the robot. 
This is based on~LineRides~\cite{rho2026linerides}. 
Each policy is using the same guideline pipeline (a series of key orientations and lines to follow and track) 
but each is trained separately on a different guideline reference 
generated via a cubic Hermite curve. 

Finally, 
the \emph{Flip}~$\pi_f$ and \emph{Lateral Jumps}~$\pi_{lj}$ policies
in~\tref{tab_robot_stunts_taxonomy}(E) 
are trained with \emph{Motion Imitation} via~\gls{imi}~\cite{kim2026flip}.
The objective is to track a series of~$M$ full robot states~$\in\Rnum^{N\times M}$, 
including all joint angles and the body pose and twist. 
With the highest dimensionality and step-wise updates, 
this leaves the narrowest margin for deviation, 
forcing the agent to mimic the reference with high accuracy. 

These task types can be deployed together 
as coordinated by the \emph{Orchestrator}, which decides
where and when each policy runs; 
examples are listed in~\tref{tab_robot_stunts_taxonomy}(F). 
All stunts in this section use the~\emph{physically locked} system, except those
in~\tref{tab_robot_stunts_taxonomy}(6, 7, 11, 15, 22--25)
which utilize the out-of-sagittal-plane \emph{Yoll} joint movement. 
For consistency, 
we omit the analysis of the stunts that use the full system
with the Yoll joint
in the main article and detail them in~\sref{sec_demo2}.

\subsection*{Reinforcement Learning (RL) Problem Formulation}
The~\gls{rl} policy optimization problem is formulated as a~\gls{cmdp}
defined by the tuple $(\mathcal{S}, \mathcal{A}, \mathcal{P}, r, \mathcal{C}, \gamma)$, 
where $\mathcal{S}$ is the state space, 
$\mathcal{A}$ is the action space, 
$\mathcal{P}(s_{t+1}|s_t, a_t)$ is the state transition probability to state~$s_{t+1}$ 
from state~$s_t$ after taking action~$a_t$, 
$r(s_t, a_t)$ is the reward function, 
$\mathcal{C} = \{c_1, ..., c_k\}$ is a set of $k$ constraint functions, and $\gamma \in [0, 1)$ is the discount factor.
The objective then is to 
find the optimal policy~$\pi^*$ that maximizes 
the expected discounted sum of future rewards 
while satisfying a set of constraints~$\mathcal{C}$. 
These constraints $c_i(s_t, a_t)~\leq~0$, 
represent the physical and operational limits of the robot.
The full objective is:
\begin{equation*}
\pi^* = \arg\max_{\pi} \mathbb{E}_{\xi \sim \pi} \left[ \sum_{t=0}^{T} \gamma^t r(s_t, a_t) \right]
\end{equation*}
\begin{equation}
\text{subject to } \quad c_i(s_t, a_t) \leq 0, \quad \forall i \in \{1,...,k\}, \forall t .
\end{equation}

Using constraints-as-terminations~\cite{chane2024cat}, 
constraint violations are handled by immediately terminating the episode, 
transforming the~\gls{cmdp}
into a standard~\acrshort{mdp}
solvable with conventional~\gls{rl} methods. 
Because the reward provides a positive average reward at each timestep, 
policies that violate constraints incur shorter episodes and lower expected returns, 
encouraging safe behavior. 
We employ~\gls{ppo}~\cite{nikita2021learning, schulman2017proximal} to solve this problem. 
Both the actor and critic are~\gls{mlp}s with three hidden layers and ELU activations.

\subsection*{Simulation and Control Loop}
We use Isaac Lab as our training environment~\cite{mittal2025isaac}.
The training iteration numbers needed for every policy are in~\tref{table_other_hyper_params}.
The policies operate at~\unit[50]{Hz}, outputting joint PD targets. 
These targets are tracked by a low-level PD controller running at~\unit[200]{Hz} in simulation, 
and at~\unit[8]{kHz} on the robot.

\subsection*{Joint and Action Spaces}
The joint space of the policies that use the~\emph{physically locked} version of~\gls{umv}
is~$q = [q_h, q_n, q_s, q_w]^T \in \Rnum^{4}$ (see~\sref{sec_umv_sup}).
For joint~$j$, the final desired torque~$\tau_j$ is defined as
\begin{equation}
\tau_j = k_\mathrm{j}^\mathrm{p}(q_{\mathrm{j}}^{\mathrm{des}}- q_\mathrm{j}) + k_\mathrm{j}^\mathrm{d}(\dot{q}_{\mathrm{j}}^{\mathrm{des}} - \dot{q}_\mathrm{j}), 
\end{equation}
where~$q_{\mathrm{j}}^{\mathrm{des}}$ 
and~$\dot{q}_{\mathrm{j}}^{\mathrm{des}}$ are the desired position 
and velocity setpoints for every joint, respectively. 
The setpoints are scaled 
versions of the actions
meaning~$q_{\mathrm{j}}^{\mathrm{des}} = k_\mathrm{j}^\mathrm{a} \cdot a_\mathrm{j}$ 
or~$\dot{q}_{\mathrm{j}}^{\mathrm{des}} = k_\mathrm{j}^\mathrm{a} \cdot a_\mathrm{j}$.
The steering~$a_s$, head~$a_h$, and neck~$a_n$ actions are defined 
as scaled position setpoints~$q_{\mathrm{j}}^{\mathrm{des}} = k_\mathrm{j}^\mathrm{a} \cdot a_\mathrm{j}$ 
while the rear wheel action~$a_w$ is defined 
as scaled velocity setpoint~$\dot{q}_{\mathrm{j}}^{\mathrm{des}} = k_\mathrm{j}^\mathrm{a} \cdot a_\mathrm{j}$. 
We chose the rear wheel action as a velocity setpoint and not position since the latter is unbounded. 
Details on the actions and the actions parameters are in~Supplementary Tables~\ref{table_obs_act}(D)~and~\ref{table_act}.

As shown in~\tref{table_act}, 
the PD gains depend on the task complexity. 
Waypoint following and driving use nominal PD gains. 
The pose reaching tasks have higher action scales on the head and neck joints, 
which we believe helps explore the explosive transition 
and generate the centroidal momentum needed to recover from a fully inverted pose. 
Conversely, 
guideline tracking and motion imitation use the highest joint stiffness 
and lowest damping, 
which we believe helps 
adhere to the reference closely with minimal deviation.

\subsection*{Observations}
Table~\ref{table_obs_act} details the observations for each of the five training types 
and the noise distributions used for sim-to-real. 
All policies share a common observation set~(\tref{table_obs_act}(A)): 
bike angular velocity~($\omega$),       
projected gravity~($g$),                
steering joint position~($q_s$),        
steering joint velocity~($\dot{q}_s$),  
rear wheel joint velocity~($\dot{q}_w$),
head joint position~($q_h$),            
head joint velocity~($\dot{q}_h$),      
neck joint position~($q_n$),            
neck joint velocity~($\dot{q}_n$), and      
last actions~($a_{t-1}$).               
Each policy adds policy-specific observations~(\tref{table_obs_act}(B)): 
waypoint following adds 
a 2D goal position command ($x_{\text{goal}}$),
a bike height ($z_{\text{bike}}$), and           
a heightmap ($H$); 
pose reaching adds the stunt trigger command~($c_{\text{stunt}}$); 
SE2 twist tracking adds the SE2 velocity command ($v_c$); 
guided tracking adds the SE2 velocity command ($v_c$), 
the stunt trigger command~($c_{\text{stunt}}$), 
the bike height ($z_{\text{bike}}$), and           
the bike global position~($x_{\text{bike}}$); 
and the imitation policy adds the phase variable~$\phi$, 
which increases linearly from 0 to 1 over the reference trajectory. 
Note that the guided tracking and motion imitation policies 
require neither the guidelines nor the references as observations. 
The references are not part of the observations and they are only used to compute the tracking rewards. 
This is because any guided tracking or motion imitation policy
is trained to track a single guideline or reference, and thus,
either a phase clock variable or the robot position is sufficient. 

To stabilize the optimization process and help the model converge faster, 
we normalize all input observations using a running empirical average. 
Furthermore, to improve the robustness and reliability of our policies,
we use an asymmetric actor-critic 
architecture~\cite{pinto2017asymmetric, abdolhosseini2019learning, hoeller2024anymal}. 
During simulation, the critic is given privileged information, 
extra environmental data and noise-free observations unavailable in the real world, 
letting it learn a much more accurate value function. 
The policy network, however, 
is restricted to the raw onboard observations available during physical deployment. 
The privileged information for each training type is shown in~\tref{table_obs_act}(C).

\subsection*{Rewards}
The reward terms for the waypoint following, pose reaching, SE2 twist tracking, guided tracking, and imitation
training types are detailed in Supplementary Tables~\ref{table_jump_reward_functions}-\ref{table_flip_reward_functions}, 
including the full list of rewards, their expressions, weights, and explanations. 
Each reward term uses the 
\textit{exponential kernel}, 
\textit{squared exponential kernel}, 
or \textit{squared} 
functions defined below. 
\begin{equation*}
r = \exp\left( - \frac{\left\|x - x^c \right\|}{\sigma^2} \right), \ \ \ 
r = \exp\left( - \frac{\left\| x - x^c \right\|^2}{\sigma^2} \right), 
\end{equation*}
\begin{equation}
r = \left\|x - x^c \right\|^2.
\label{eq:squared_function}
\end{equation}
The variable~$x$ is the input value, $x^c$ is the reference, and $\sigma$ is the standard deviation.
We broadly categorize the rewards into three types: \emph{Task}, \emph{Style}, and \emph{Regularization} Rewards.

\emph{Task Rewards} are the main terms that encourage the robot to accomplish the task. 
For SE2 twist tracking, they minimize the linear and angular velocities tracking errors. 
For waypoint following, such as the perceptive jump, 
one reward minimizes the Euclidean distance between the robot and the goal, 
and another encourages the robot to move toward the goal. 
For complex acrobatics, 
the imitation policy optimizes 
a dense reference-tracking objective 
that minimizes the error between 
the bike base and joint states against a target reference trajectory. 
For pose reaching, the task rewards are conditioned on the stunt trigger. 
In the kip up policy, for instance, 
an inactive trigger encourages the robot to remain upside down~(\tref{table_kip_reward_functions}(A)), 
while an active trigger encourages it to reach the upright pose~(\tref{table_kip_reward_functions}(B)). 
For motion imitation, the task rewards are conditioned on task completion: 
the imitation rewards in~\tref{table_flip_reward_functions}(A) are active until 
the robot tracks all references (\ie the phase clock is 1), 
after which the robot is encouraged to stand straight, 
track a certain joint position, and reduce jitter~(\tref{table_flip_reward_functions}(B)). 

\emph{Style Rewards} are secondary terms encouraging a certain aesthetic behavior or style. 
For example, waypoint reaching has a reward discouraging fast motion, 
and driving, jumping, and guideline policies use a base orientation penalty to prevent 
excessive leaning or rolling from vertical. 

\emph{Regularization Rewards} 
reduce the sim-to-real gap by avoiding overfitting to the training data 
and, more importantly, prevent the robot from reaching its actuation and physical limits. 
One reward common to all training types encourages the robot to minimize action 
spikes for smoother, less jerky behavior, 
and separate losses penalize extreme joint torques, excessive link velocities, 
and mechanical joint chattering. 
Stunt-specific regularization terms are introduced based on danger profiles: 
guideline tracking and motion imitation apply contact force penalties 
to suppress damaging landing impacts, 
whereas waypoint following uses a mechanical power penalty~($\Sigma \tau \dot{q}$) 
to minimize electrical power surges during steady-state operation.

The total reward is the weighted sum of all three categories:
\begin{equation}
r_\text{total} = \sum_{i} w_i \  r_{\text{task}, i} \ + \ \sum_j w_j \ r_{\text{style}, j} \ + \ \sum_k w_k \ r_{\text{reg.}, k}.
\end{equation}

\subsection*{Termination Criteria}
All the termination terms are detailed in~\tref{table_terminations_all}. 
An episode terminates instantly 
if a joint hits its physical position or velocity limits, 
if the rear wheel passes a safe velocity threshold, 
if a motor exceeds its maximum torque, velocity, or power rating, 
or if the robot falls or collides with the ground unexpectedly. 
For the guided tracking and motion imitation tasks, 
we also terminate if the robot deviates from the references, 
and for most policies we terminate if the wheels hit the ground hard.

\subsection*{Curriculum Learning}
We introduce curriculum learning in three ways. 
First, 
and only for waypoint following, 
we use the terrain-based curriculum described earlier: 
terrain difficulty increases if an episode ends with the robot close to or at the goal, 
and decreases otherwise. 
Second, 
we apply a curriculum during reward computation. 
For waypoint following, pose reaching, and SE2 twist tracking, 
we modify specific reward weights depending on the iteration number 
(Tables~\ref{table_jump_reward_functions}-\ref{table_drive_reward_functions}), 
while for guided tracking and motion imitation 
we modify specific reward weights after the robot succeeds in tracking the references 
(Tables~\ref{table_guideline_reward_functions},\ref{table_flip_reward_functions}).
Third, 
we apply a curriculum during termination computation
as shown in~\tref{table_curr}.
This curriculum activates the termination terms 
either after a certain number of iterations as for the waypoint following, pose reaching and motion imitation tasks, 
or after a certain number of iterations after the robot succeeds in tracking the references 
as for the guided tracking tasks. 

\subsection*{Training Algorithm}
The training hyperparameters in~Tables~\ref{table_common_hyper_params} and~\ref{table_other_hyper_params} 
are mostly common across tasks except for
the number of iterations, network size, and entropy coefficient. 
One important finding is that
the entropy coefficient varies inversely with task complexity. 
The waypoint following task, of lowest complexity, 
needed the highest entropy coefficient: 
the lower the complexity, the more we need to encourage exploration of different actions. 
Conversely, motion imitation is high complexity, 
so rather than exploring we want the policy to tightly track the reference, 
requiring a smaller entropy coefficient. 
The iteration count also varies inversely with complexity: 
the waypoint following tasks train in 30K iterations 
while motion imitation trains in 15K, 
consistent with the robot needing more time to learn an unguided emergent jump 
than an imitated flip.

\subsection*{Sim-to-Real}
We utilize multiple techniques to bridge the sim-to-real gap
from several related work including but not limited
to~\cite{tobin2017domain, peng2018sim, hwangbo2019learning, tan2018sim, choi2023learning}.

\paragraph*{Observation Noise} 
As shown in~\tref{table_obs_act}, 
realistic sensor noise is modeled as uniform or Gaussian distributions and injected into 
the policy observations during training 
to account for onboard encoder and inertial measurement unit variations. 
To account for state estimation and perception errors, 
we also inject random noise and drift into the heightmap~(\tref{table_domain_randomization_complete}(E)).

\paragraph*{Domain Randomization}
As shown in~\tref{table_domain_randomization_complete}(A),
physical parameters including 
ground friction coefficients, 
link masses, 
center of mass coordinates of each link, 
and motor strength are continuously randomized between training episodes. 
Actuator imperfections between robots
are modeled by randomizing the PD gains and motor joint friction,
and by introducing communication delays. 
Random external disturbances are applied directly to the bike base, 
as shown in~\tref{table_domain_randomization_complete}(D).

\paragraph*{State Initialization}
The initial conditions of each training episode are randomized to maximize exploration 
(\tref{table_domain_randomization_complete}(B, C)). 
The waypoint following, SE2 twist, 
and guided tracking tasks initialize the robot with typical pose variations 
encountered during deployment, 
whereas pose reaching uses a targeted initialization, 
e.g., the kip up policy spawns the robot in a static, fully inverted orientation, as in the real world. 
For motion imitation, 
\gls{rsi}~\cite{kim2026flip, peng2018deepmimic} 
samples initial poses directly from the reference trajectories: 
50\%~of episodes begin from an initial state and the remaining 50\% 
from random states sampled along the reference, 
exposing the policy to critical states early in training~(\tref{table_domain_randomization_complete}(F)). 
For waypoint following, 
we also randomize the location and height of the ring (table) and the gap between two rings
(\tref{table_domain_randomization_complete}(E)).

\paragraph*{Wheels Modeling}
The compliant wheels constantly interact with the tables and ground, 
yet most simulators are tuned for legged robots, 
making accurate wheel modeling both important and difficult. 
We modeled each wheel as a compliant body parameterized by stiffness and damping. 
We selected the contact parameters and geometry that best matched drop-test data on the physical robot. 
Full details are given in~\sref{sec_wheels}.

\paragraph*{Actuators Modeling}
To better match the physical actuators, 
the simulation incorporates realistic torque–speed characteristics, 
with joint torque limits computed dynamically from the motors' electromechanical constraints 
rather than imposed as fixed saturation bounds. 
The torque constants and nominal PD gains in~\tref{table_act} 
were experimentally identified on the physical robot. 
Full details are given in~\sref{sec_actuators}.

\subsection*{Hardware Deployment}
The hardware deployment pipeline is built in C++. 
Before testing on the real robot, 
we evaluate the policy in Isaac Lab~\cite{mittal2025isaac}, 
the same simulator used for training.
We then test the policy on the deployment code in MuJoCo~\cite{todorov2012mujoco}, since it is a different simulator,
which allows us to ensure that the policy did not overfit to the simulation environment it was trained with. 
The policies are exported in an ONNX~\cite{onnx} format
using an exporter similar to~\cite{exploy2026}. 
Controls and state estimation are running on an UpExtreme i12~\cite{upex}. 
When deploying on the real robot, the policies require a state estimator 
to construct their observations from the onboard sensors. 
We detail this state-estimation pipeline in~\sref{sec_state_est}, 
and refer the reader to~\cite{RAIInstitute2026UMV} for more information.

\subsection*{Orchestrator}
\begin{figure*}
\centering
\includegraphics[width=0.99\textwidth]{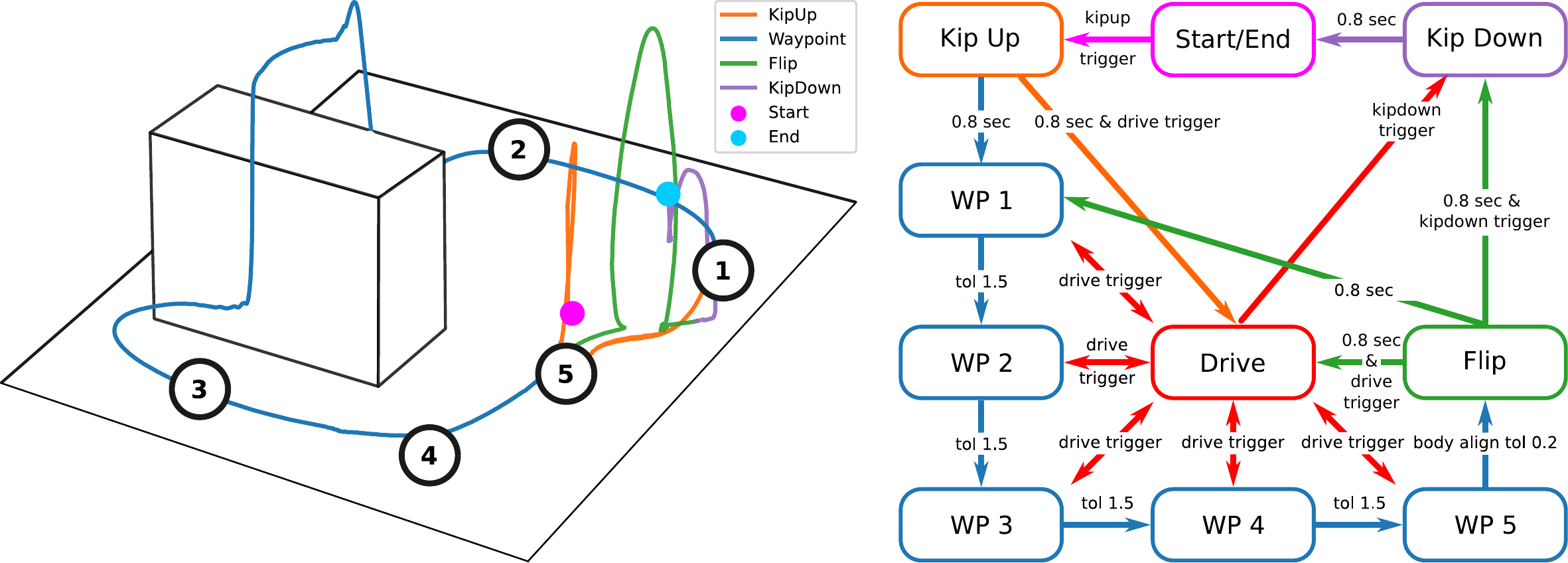}
\caption{\textbf{Autonomous execution of the combined kip, jump and flip stunt coordinated by the orchestrator.}
(\textbf{A}) A single typical hardware experiment of this stunt showing a kip-up (orange), 
jump/waypoint-following (blue), flip (green), and kip-down (purple). 
Numbered circles denote waypoint locations, and the magenta and cyan markers indicate the start and end positions, respectively.
(\textbf{B}) The orchestrator diagram governing this stunt.}

\label{fig_orch_odnt}
\end{figure*}

The individual stunts~(flips, kip-ups, jumps, etc.,) shown in~\tref{tab_robot_stunts_taxonomy}(A)--(E), 
are independent~\gls{rl} policies that each perform a certain task.
The orchestrator determines both 
the execution order of these policies and the conditions under which transitions occur. 
Importantly, it should not transition between behaviors merely on a button press, 
but decide to transition in a dynamic and smooth way. 
For example, 
the flip \& drive stunt~(\tref{tab_robot_stunts_taxonomy}(17))
is a sequence of flips and driving policies, 
and the robot needs to know when to transition between successive flips 
and between a flip and driving. 
To transition from a flip to driving, the robot must first be in a safe state, 
ensuring the flip succeeded rather than switching while airborne, 
even if the orchestrator is commanded to drive. 
We designed the orchestrator as an~\gls{fsm} based on sequential composition~\cite{burridge1999sequential}, 
which defines state-dependent, rather than purely logical, relationships between policies.
We detail this design choice in~\sref{sec_orch}.

The orchestrator consists of two components:~\emph{Nodes} and \emph{Edges}. 
A~\emph{Node} is an~\gls{rl} policy, i.e., a state of the~\gls{fsm}, 
and an~\emph{Edge} is the transition condition to move from one node to another, 
where a node may have more than one edge. 
In other words, each node corresponds to a policy, 
while each directed edge represents a validated transition between policies.

\fref{fig_orch_odnt}(B) shows the orchestrator diagram for 
the kip, jump \& flip stunt in~\tref{tab_robot_stunts_taxonomy}(21). 
\fref{fig_orch_odnt}(A) shows the typical trajectory the robot takes and the waypoints that are provided
during such stunts.
The robot always starts and ends in the upside-down tucked configuration, 
where the orchestrator sits in an idle state. 
Once a kip up trigger is pressed, it switches to the kip up policy.
Then, autonomously after~\unit[0.8]{s}, it transitions to the jump/waypoint-following policy 
and drives the robot from~WP1 through~WP5. 
Across these waypoints the same policy runs throughout; 
the orchestrator only selects which waypoint is active, 
advancing to the next one once the robot is within a~\unit[1.5]{m} radius tolerance. 
At~WP5, the orchestrator prepares a flip but triggers it only once the robot reaches~WP5 
and its body yaw is aligned parallel to the table within~\unit[0.2]{rad}, 
so it does not strike the table. 
After~\unit[0.8]{s} from the flip, the robot can transition to one of three states. 
Autonomously,
the orchestrator drives it back to~WP1 to repeat the loop.
If the kip down trigger is pressed, 
it kips down to the upside-down position in~\unit[0.8]{s} from which the sequence can restart with a kip up,
or a if a drive trigger is pressed 
the orchestrator will transition to the drive policy where
the robot accepts inputs from the joystick. 
Overriding all of these, a safety trigger lets a user take control through the drive state at any time, 
which can also act as a track-stand state that simply holds the robot in place. 
\sref{sec_orch} and~\fref{fig_orch_odnt_supp}
detail the orchestrator diagrams used for the other stunts in this work.

\renewcommand{\refname}{References and Notes}
\bibliography{references.bib}

\paragraph*{Acknowledgments}
We thank the entire hardware, simulation, machine learning, and controls team of~\gls{umv} including
A.~Preston, A.~Bahner,  A.~Agarwal,  A.~Wollschläger,  A.~Erdheim,  B.~Bokser,  D.~Gonzalez,  D.~Perry,  E.~Lin,  
F.~Jenelten,  F.~Yu,  F.~Gillespie,  H.~Mayne,  H.~Hansen,  I.~Tinman,  J.~Smith,  J.~Tigue,  J.~Preisig,  
K.~Sharma,  K.~Ang,  L.~Margolese,  M.~Wang,  M.~Zenti,  M.~Pickett,  N.~Lin,  O.~Frias,  R.~Burnham, 
S.~Biddlestone,   S.~Mayorga,  S.~Singh,  T.~McCollum,  V.~Dimitrov,  and W.~Moyne for their 
dedicated support throughout the development and testing of the \gls{umv} platform. 
We also acknowledge the excellent interns who contributed to the UMV project: 
A.~Tajbakhsh,  J.~Kim,  R.~Grummer,  and S.~Rho.  
We sincerely thank the reviewers and editorial team for their insightful feedback and guidance, 
which greatly improved the quality of this manuscript.
\textbf{Funding:}
This research is supported by the Robotics and AI Institute, 
which, in turn, receives funding from Hyundai Motor Group. 
The opinions expressed herein are those of the authors and independent of the views of the funding organizations. 
\textbf{Author Contributors:}
The Contributors
developed the embedded and training-tool software, 
trained the policies, 
and oversaw the hardware experiments on robots. 
They also contributed to the writing of the manuscript. 
They assisted with engineering tasks to support the project,
including deployment code on hardware, 
data collection and curation, 
as well as supporting hardware experiments.  
The Project Leads set the overarching vision by defining the core scientific goals and managed the project. 
\textbf{Competing interests:}
There are no competing interests to declare.
\textbf{Data, code, and materials availability:}
Data to support the paper's conclusions are present herein or in the Supplementary Materials.
The data and figure generation code for this study will be available if approved by our internal review process.
We note that while all ideas, methods, and results are original to the authors, 
AI-based tools were used solely to refine the clarity and readability of the manuscript text. 

\section*{Supplementary Materials}
\resizebox{\columnwidth}{!}{
\begin{tabular}{cl}
Section S1.  & Acronyms\\
Section S2.  & Symbols\\
Section S3.  & Details on the UMV Robot \\
Section S4.  & Ablation Studies\\
Section S5.  & Training Details\\
Section S6.  & Wheels Modeling\\
Section S7.  & Actuators Modeling\\
Section S8.  & State Estimation\\
Section S9.  & Orchestrator Details\\
~&~\\
Figure S1.   & UMV Versions \& Representations\\
Figure S2.   & Ablation Studies\\
Figure S3.   & Training Setup\\
Figure S4.   & Additional Orchestrator Diagrams\\
~&~\\
Table S1.    & Observations \& Actions\\
Table S2.    & Actions Gains\\
Table S3.    & Waypoint Following Jump Rewards\\
Table S4.    & Pose Reaching Kip Up \& Down Rewards\\
Table S5.    & SE2 Twist Tracking Drive Rewards\\
Table S6.    & Guided Tracking Rewards\\
Table S7.    & Motion Imitation Rewards\\
Table S8.    & Domain Randomization\\
Table S9.    & Termination Criteria\\
Table S10.   & Curriculum Learning\\
Table S11.   & Common Training Algorithm Params\\
Table S12.   & Task Specific Training Algorithm Params\\
~&~\\
Movie S1.    & Results \& Methods Overview\\
Movie S2.    & Kips, Jumps \& Flips\\
Movie S3.    & Lateral Jumps \& Wheelies\\
Movie S4.    & Jumps\\
Movie S5.    & Kips \& Drive\\
Movie S6.    & Motion Imitation: Flips\\
Movie S7.    & Guided Tracking: Three-point Turns \& Bunny Hops\\
Movie S8.    & Park From Anywhere\\
Movie S9.    & Driving\\
\end{tabular}
}

\clearpage\newpage
\renewcommand{\thefigure}{S\arabic{figure}}
\renewcommand{\thetable}{S\arabic{table}}
\renewcommand{\theequation}{S\arabic{equation}}
\renewcommand{\thepage}{S\arabic{page}}
\renewcommand{\thesection}{S\arabic{section}}

\makeatletter
\renewcommand{\@seccntformat}[1]{%
  \ifnum\pdfstrcmp{#1}{section}=0
    Section \csname the#1\endcsname.\quad
  \else
    \csname the#1\endcsname.\quad
  \fi
}
\makeatother

\setcounter{figure}{0}
\setcounter{table}{0}
\setcounter{equation}{0}
\setcounter{page}{1}
\setcounter{section}{0}
\onecolumn

\noindent{\huge \bfseries Supplementary Materials - Acrobatic Bicycle Stunts via Reinforcement Learning\par}

\printglossary[type=\acronymtype,title=Acronyms]

\section{Symbols}\label{sec_nom}

\subsection*{Robot Model and Joints}
\begin{tabular}{cl}
$A_i$ & Actuators ($A_0$ through $A_5$) \\
$q_h$ & Head joint angle (between Head and Neck links) \\
$q_l$ & Left Fin (arm) joint angle \\
$q_r$ & Right Fin (arm) joint angle \\
$q_s$ & Steering joint angle \\
$q_w$ & Rear wheel joint angle \\
$q_n$ & Neck joint angle (passive, serial representation) \\
$q_y$ & Yoll (out-of-plane) joint angle (passive) \\
$q_f$ & Front wheel joint angle (passive, unactuated) \\
$q$ & Joint positions vector \\
$\dot{q}$ & Joint velocities vector \\
$q_\text{des}$ & Desired joint positions vector \\
$n$ & Total number of actuated DoFs \\
\end{tabular}

\subsection*{Control and Actuation}
\begin{tabular}{cl}
$\tau_j$ & Desired torque for joint $j$ \\
$k_j^\mathrm{p},\ k_j^\mathrm{d}$ & PD position and velocity gains for joint $j$ \\
$k_j^\mathrm{a}$ & Action scaling factor for joint $j$ \\
$q_j^\text{des},\ \dot{q}_j^\text{des}$ & Desired position/velocity setpoints for joint $j$ \\
$a_j$ & Action for joint $j$ (e.g., $a_s$, $a_h$, $a_n$, $a_w$) \\
$a_{t-1}$ & Last actions \\
\end{tabular}

\subsection*{Observations and Commands}
\begin{tabular}{cl}
$\omega$ & Angular velocity of the bike frame \\
$g$ & Projected gravity \\
$x_\text{goal}$ & 2D goal position command \\
$z_\text{bike}$ & Bike height \\
$x_\text{bike}$ & Bike global position \\
$H$ & Heightmap \\
$c_\text{stunt}$ & Stunt trigger command \\
$v_c$ & SE2 velocity (twist) command \\
$v_c^x$ & Commanded forward velocity \\
$\omega_c^z$ & Commanded yaw rate \\
$\phi$ & Imitation phase variable (increases from 0 to 1) \\
\end{tabular}

\subsection*{Reward Terms}
\begin{tabular}{cl}
$x$ & Input value to a reward kernel \\
$x^c$ & Reference value \\
$\sigma$ & Standard deviation \\
$r_\text{total}$ & Total (weighted) reward \\
$w_i,\ w_j,\ w_k$ & Task, style, and regularization reward weights \\
$M$ & Number of reference poses/keyframes \\
$N$ & Full robot state dimension \\
\end{tabular}

\subsection*{Policies}
\begin{tabular}{cl}
$\pi_j$ & Jump (waypoint following) policy \\
$\pi_{mj}$ & Multi-table jump policy \\
$\pi_{ku},\ \pi_{kd}$ & Kip up and kip down policies \\
$\pi_d$ & Driving (SE2 twist) policy \\
$\pi_h,\ \pi_{tpt}$ & Bunny hops and three-point-turn policies \\
$\pi_f$ & Flips (motion imitation) policy \\
$\pi_w$ & Wheelie policy \\
$\pi_{aw},\ \pi_{sw}$ & Autonomous (waypoint-following) and steerable wheelie policies \\
$\pi_{lj},\ \pi_{lj0}$ & Imitated and waypoint-following lateral jump policies \\
\end{tabular}

\section{Details on the UMV Robot}\label{sec_umv_sup}
The~\gls{umv} robot, shown in~\fref{fig_umv_overview}, 
is a custom-built bicycle platform designed for high-performance and agility.
It weighs \unit[23.5]{kg} and stands \unit[0.8]{m} tall when fully crouched. 
UMV
has eleven links: 
Head, 
left and right Fins, 
Neck, 
left and right Tie Rods, 
Bike, 
Fork, 
out-of-plane Yoll, 
and front and rear Wheels.
\gls{umv} has five actuated~\gls{dofs}:
the~\emph{Head}, 
\emph{Left Fin}, \emph{Right Fin}, 
\emph{Steering}, and \emph{Rear Wheel} joints, respectively.
Forward motion comes from the rear wheel drive~$q_w$,
and steering from~$q_s$. 
The remaining three~\gls{dofs}, $q_h$, $q_l$, and $q_r$, 
control the position and orientation of the \emph{Head} 
relative to the \emph{Bike} through a spatial linkage 
consisting of the \emph{Neck} link and left and right \emph{Tie Rods}. 
The five~\gls{dofs} are driven by six actuators, $A_0$ through $A_5$.
Actuators $A_2$ and $A_3$ are coupled to move $q_h$ directly, 
representing the angle between the \emph{Neck} and \emph{Head}. 
Actuators $A_0$ ($q_r$) and $A_1$ ($q_l$) are coupled through the left and right \emph{Tie Rods}, respectively,
and work together 
with actuators $A_2$ and $A_3$
to change the orientation of the \emph{Head} relative to the \emph{Bike}.
Consequently, differential motion between $q_l$ and $q_r$ produces out-of-sagittal-plane movement 
that we refer to as the \emph{Yoll} motion
since the joint partly yaws and partly rolls. 
The front wheel's joint~$q_f$ is passive and unactuated.

The roll-cage-protected \emph{Head} houses the jumping actuators, 
power source, and main computer. 
Wiring passes through the hollow \emph{Neck} to the \emph{Bike} link. 
The \emph{Bike} mechanism is based on a carbon-fiber commercial children's push bike 
(Specialized Hotwalk Carbon~\cite{specialized_hotwalk_carbon}) and integrates 
custom steering, drive, and power-electronics housings, with structural reinforcement 
at the rear wheel for dynamic takeoff and landing.
The onboard computer is an UP Xtreme i12~\cite{upex} running a \unit[1]{kHz} control loop, 
while motor drivers run at \unit[8]{kHz} and close a PD loop on position and velocity setpoints.
For a detailed explanation of 
the system design of~\gls{umv}
and the different kinematic representations,  
we refer the reader to 
Bokser~\etal~\cite{RAIInstitute2026UMV}. 

\begin{figure*}
\centering
\includegraphics[width=\textwidth]{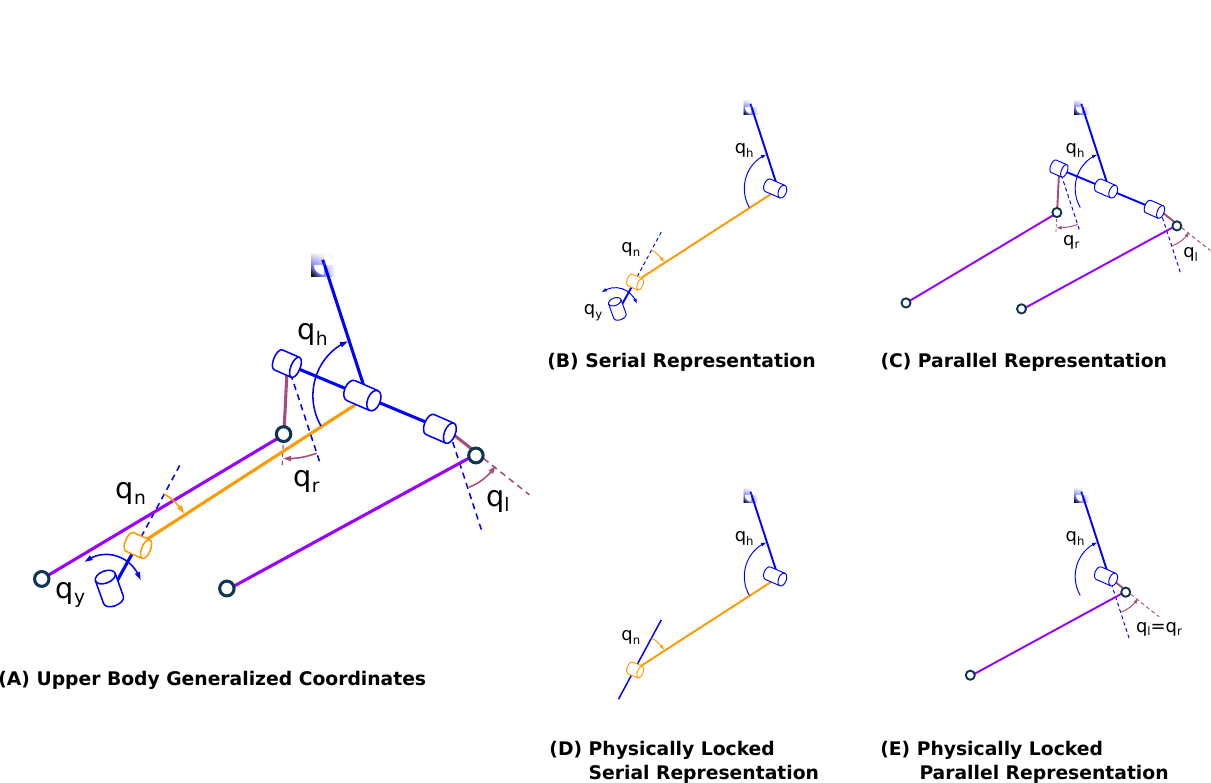}
\caption{\textbf{UMV Versions \& Representations.} 
(\textbf{A}) the full representation and generalized coordinates of~\gls{umv}.
\gls{umv}~has two different versions, one with the Yoll joint~(\textbf{B},\textbf{C}), 
and one with the Yoll joint physically locked~(\textbf{D},\textbf{E}). 
\gls{umv}~also has two different kinematic representations denoted by the serial~(\textbf{B},\textbf{D}) 
and parallel representations~(\textbf{C},\textbf{E}).
}
\label{fig_umv_sup}
\end{figure*}

There are two versions of~\gls{umv}: one with~the \emph{Yoll} and one without. 
We refer to the one without the \emph{Yoll} joint as \emph{physically locked}.
All the stunts and policies explained in~\tref{tab_robot_stunts_taxonomy}
are using the physically locked version except 
those in~\tref{tab_robot_stunts_taxonomy}(6, 7, 11, 15, 22--25).
There are two kinematic representations of~\gls{umv}:serial and parallel representations. 
The parallel representation is the one that includes the spatial linkages between the fins, tie rods, and the neck.
The serial representation assumes that the neck, and yoll can be directly controlled. 
The generalized coordinates of~\gls{umv} are defined by
$q = [ q_h, q_l, q_r, q_n, q_y, q_s, q_w ]^T$ $\in\Rnum^7$.
The physically locked system assumes $q_r = q_l$ yielding $q_y = 0$. 
All the aforementioned representations and topologies are illustrated in~\fref{fig_umv_sup}.

\section{Ablation Studies}\label{sec_ablations}
\begin{figure*}
\centering
\includegraphics[width=\textwidth]{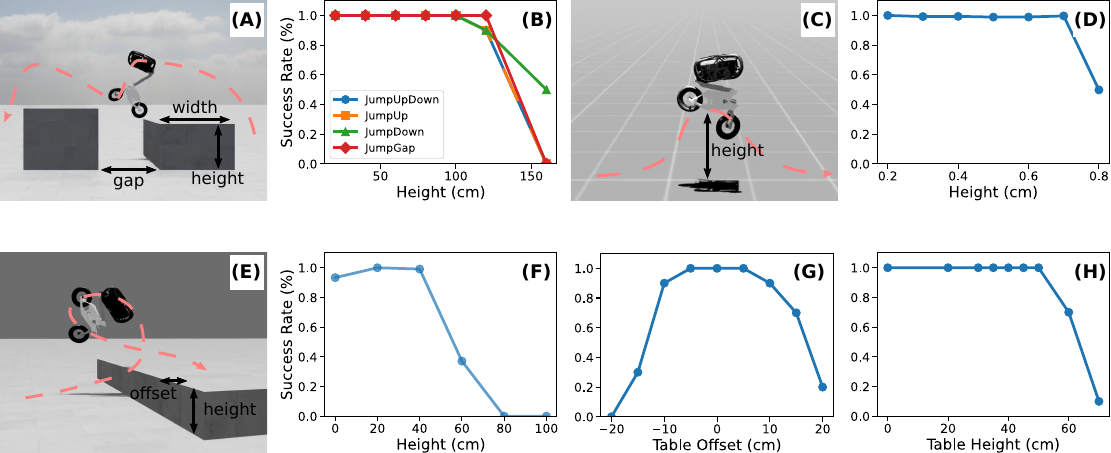}
\caption{\textbf{Ablation Studies.} 
Several simulation~(\textbf{B},\textbf{G},\textbf{H}) and training~(\textbf{D},\textbf{F}) ablations studies to test the operational limits 
of the stunts presented in this work. 
(\textbf{A,B})~Success rate of varying four jumping variants 
(\textit{JumpUpDown}, \textit{JumpUp}, \textit{JumpDown}, \textit{JumpGap}) versus the parameter height
using the jump policy. 
(\textbf{C,D})~Success rate of varying the hop height during training guided tracking hop policy. 
(\textbf{E--H})~Ablation studies on the flip on table policy 
during training~(\textbf{F}), and 
in simulation~(\textbf{G},\textbf{H}). 
Success rate of 
varying the table height in training~(\textbf{F}), and
varying the table offset~(\textbf{G}) and height~(\textbf{H}) in simulation.}
\label{fig_ablations}
\end{figure*}

Here, 
we detail the ablation studies that we performed on some of our~\gls{rl} policies. 
These studies test the operational limits of the 
jump policies trained with waypoint following, 
the hop stunt trained with guided tracking, and 
the flip-on-table stunt trained with motion imitation. 
The success rates of these ablations are illustrated in~\fref{fig_ablations}.

In the first study, 
we investigate how high the robot can jump up (JumpUp), down (JumpDown), and up and down (JumpUpDown) a table, 
and how large a gap it can clear between two tables (JumpGap). 
These four tests are reported in~\fref{fig_ablations}(A,B). 
For the JumpUpDown, JumpUp, and JumpDown cases, we vary the table height over~\unit[0--100]{cm}, 
while for the JumpGap case we vary the gap width over~\unit[0--100]{cm} between two tables of~\unit[40]{cm}.
The jumping controllers demonstrate a high tolerance for varying terrain heights. 
Across all four variants (\textit{JumpUpDown}, \textit{JumpUp}, \textit{JumpDown}, and \textit{JumpGap}), 
the robot maintains a perfect \unit[100]{\%} success rate for table heights and gap widths up to \unit[100]{cm}. 
Performance begins to drop past the~\unit[120]{cm} threshold, 
with most variants hitting a hard physical limit and 
failing completely at \unit[160]{cm}. 
Note that this policy was trained with a maximum table height of~\unit[100]{cm}.
The \textit{JumpDown} policy proves to be the most resilient at the extreme bounds, 
retaining a \unit[50]{\%} success rate at a table height of \unit[160]{cm}.

In the second study, 
we investigate how high the robot can hop without degrading performance 
when training multiple guided-tracking hop policies. 
To do so, we trained seven hop policies~$\pi_h$ with the hop height varying between \unit[0.2--0.8]{m}. 
The success rates are shown in~\fref{fig_ablations}(C,D). 
The same policy and training pipeline can hop up to~\unit[70]{cm} without any deterioration in 
behavior, and reach a \unit[50]{\%} success rate at~\unit[80]{cm}. 

In the third study, we ablate the performance of the flip-on-table stunt, 
a motion-imitation stunt in which the policy is trained to flip up onto a table. 
First, we test the success rate of training policies with various table heights, 
as reported in~\fref{fig_ablations}(F). 
The same training formulation and reference can be used to perform a flip on tables up to~\unit[40]{cm} tall, 
while taller tables show a dramatic deterioration in training success. 
Then, given the~\unit[40]{cm} flip-on-table policy, we evaluate it in simulation under various table settings. 
We test the policy on tables of varying height over~\unit[0--70]{cm}, as shown in~\fref{fig_ablations}(H), 
and on a \unit[40]{cm} table while varying the table's location 
with an offset of~\unit[$\pm$20]{cm}, as shown in~\fref{fig_ablations}(G).
Unlike the standard jump, the flip-on-table policy operates within a tighter physical 
envelope due to the complex coordination required to complete a full rotation. 
As shown in~\fref{fig_ablations}(H),
the flip controller clears table heights up to \unit[50]{cm} with flawless reliability. 
Beyond \unit[50]{cm}, 
success rates decline sharply, 
falling to roughly \unit[10]{\%} at \unit[70]{cm}. 
This drop-off marks the threshold where the robot 
can no longer generate enough vertical clearance to complete its angular rotation before 
making premature contact with the elevated surface. 
For the offset test in~\fref{fig_ablations}(G),
the resulting curve forms an asymmetric plateau. 
The robot handles a tight tolerance window perfectly between~\unit[$\pm$5]{cm}. 
However, the system is significantly less forgiving of negative offsets (undershooting the table), 
where the success rate plummets to \unit[0]{\%} at \unit[-20]{cm}
due to front-wheel collisions with the face of the platform. 
Conversely, it shows greater compliance when overshooting (positive offsets), 
maintaining a \unit[70]{\%} success rate even when misplaced by \unit[+15]{cm}, 
which indicates that the underlying balance controllers can safely adapt to late landings.

\section{Training Details}\label{sec_training_jump}\label{sec_training_pose}\label{sec_demo2}

\subsection*{Training Waypoint Following Jump Policies}
Figure~\ref{fig_setup}(A) illustrates the training environment used for learning the jump policies. 
The robot is initialized on a central platform surrounded by an inner ring, a gap, and an outer ring. 
During training, a single target waypoint is sampled and remains fixed for the entire episode; unlike deployment, 
where multiple waypoints may be tracked sequentially, commands are not resampled online. 
Target locations are sampled outside the platform region, 
at distances of up to \unit[8]{m} from the robot and within a forward-facing sector of $\pm \pi/2$, 
preventing commands from being placed behind the robot. 
To improve robustness and 
expose the policy to a wider range of environments,
the dimensions of the platform, inner and outer rings, gap width, 
obstacle heights, and command sampling range are randomized across training episodes.
The randomizations are shown in~\tref{table_domain_randomization_complete}(E).

\subsection*{Training Pose Reaching Kip Up and Kip Down Policies}
\begin{figure*}
\centering
\includegraphics[width=\columnwidth]{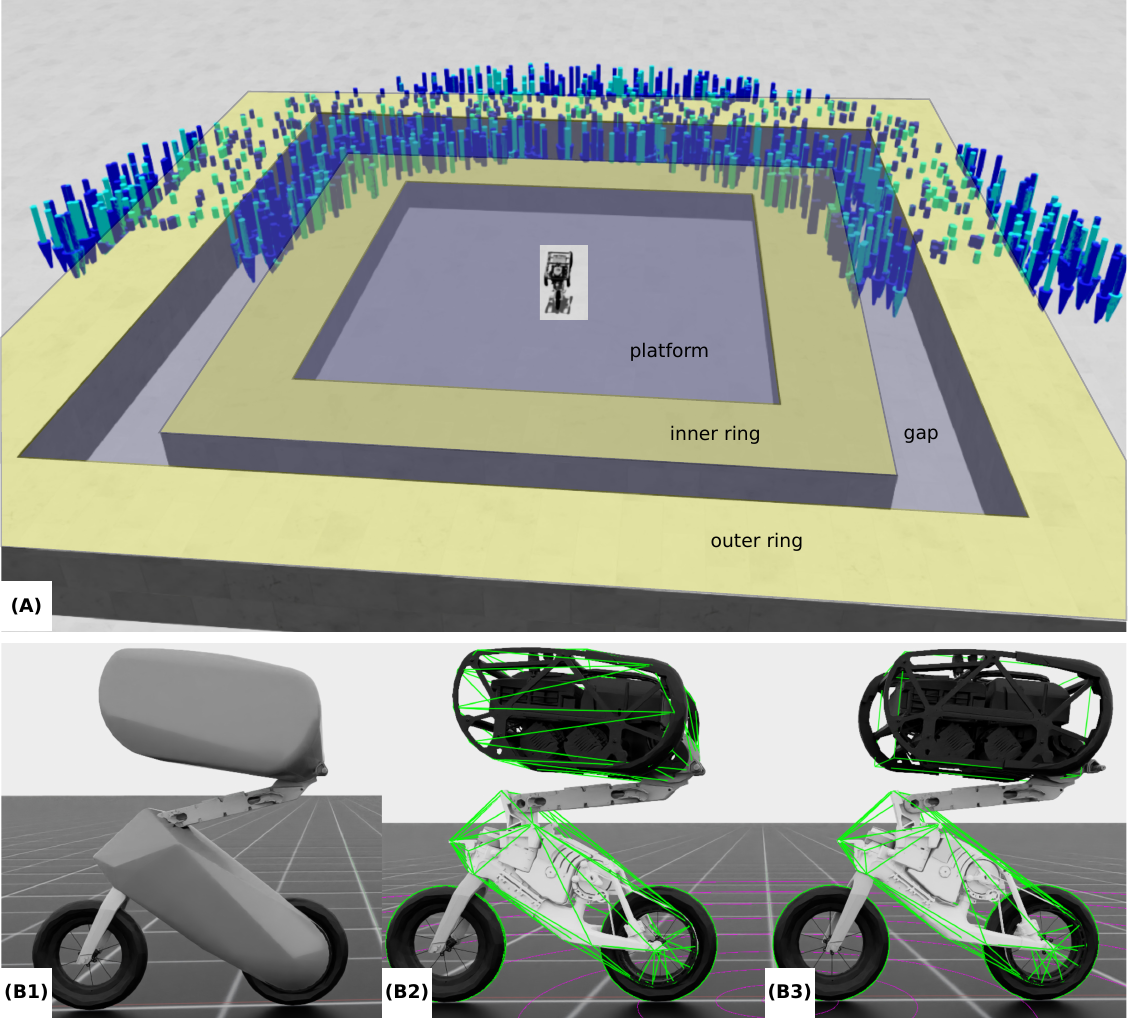}
\caption{\textbf{Training Setup.} 
(\textbf{A})~Training environment for the waypoint-following jump policies: 
the robot starts on a central platform surrounded by an inner ring, a gap, and an outer ring, 
with a single fixed target waypoint sampled each episode. 
(\textbf{B})~Collision geometry used for the kip task (\textbf{B1},\textbf{B2}) versus most other tasks (\textbf{B3}); 
both model the wheels, bike frame, and head, but the kip task uses a higher-resolution 
collision triangular-mesh head approximation rather than a simple box, 
since the kip begins with the head in contact with the ground.}
\label{fig_setup}
\end{figure*}

Figure~\ref{fig_setup}(B) compares the collision geometry used for most tasks (B3) 
with the collision geometry used for the kip task (B2). 
In both cases, the collision model captures only the wheels, 
bike frame, and head, while the neck is omitted. 
The primary difference lies in the head representation: 
the default collision model approximates the head using a simple box geometry, 
whereas the kip task employs a more accurate triangular-mesh approximation. 
This increased resolution is necessary because the kip maneuver begins with the head in direct contact with the ground, 
making the head geometry a critical factor in the contact interaction dynamics. 
For all other tasks, the simplified head model is sufficient, as the head rarely contacts the environment.

\subsection*{Training Policies with the Yoll Joint: Wheelie and Lateral Jump Policies}
The orchestrated stunts 
in~\tref{tab_robot_stunts_taxonomy}(22--25) and~\fref{fig_demo2}, 
use the version of~\gls{umv} with the unlocked~\emph{Yoll} joint. 
They rely on three types of policies: a single-wheel wheelie policy, 
an imitated lateral jump policy~$\pi_{lj}$, 
and the waypoint-following jump policy~$\pi_j$.

We train two variants of the single-wheel wheelie policy: 
an autonomous, waypoint-following variant~$\pi_{aw}$ that aligns the robot with the table on its own, 
and a steerable, joystick-commanded variant~$\pi_{sw}$ that a user drives blindly. 
Details on the wheelie policies can be found in~\cite{RAIInstitute2026UMV}.
To train the imitated lateral jump policy~$\pi_{lj}$, 
we first train a waypoint-following lateral jump policy~$\pi_{lj0}$ with task rewards only, 
then use the reference trajectories produced by this policy to train the imitated policy~$\pi_{lj}$ 
with the full reward set.
The reason we used this two-stage training instead of training a single waypoint following lateral jump policy 
was that motion imitation took much less time to train and so we were able to do sim-to-real iterations much faster. 
Training these aforementioned policies are similar to training the policies on the physically locked robot, 
the main difference being that the~\emph{Yoll} joint is added to the observations and actions. 
The rest of the~\gls{rl} problem (rewards, terminations, curricula, sim-to-real, etc.) 
follow the same formulations described in this work. 

When deploying the orchestrated stunts 
in~\tref{tab_robot_stunts_taxonomy}(22--25) and~\fref{fig_demo2},
the orchestrator was also used to orchestrate an emergency ``bail out'' strategy in which
if we notice that the robot is going to fail or fall, we trigger a transition
and switch to a policy that would stabilize the robot first, before proceeding with the stunt.

\section{Wheels Modeling}\label{sec_wheels}
The wheels on the robot are relatively compliant 
and constantly hit and interact with the tables and the ground, 
while most simulation environments are tailored for legged robots and not tuned on wheels. 
This makes modeling the physical parameters and geometry of the wheels both important and hard. 
To do so, as mentioned in~\cite{RAIInstitute2026UMV}, 
we built a testing jig to perform drop tests on the wheels in 2D, 
dropping the wheels onto a force plate so that we could measure the timeseries of the wheel height 
and the forces on the ground. 
To improve the reliability of this test, 
we emulated the robot's mass by adding weights of different values to the wheels, 
and we tested several tire pressures. 
In simulation, we implemented an identical setting 
and modeled the wheel as a compliant body parameterized by stiffness and damping parameters. 
Since the contact models in simulation are not fully disclosed, 
we avoided parametric system identification: 
we sampled many different values of the wheel's contact parameters 
and picked the ones that matched the position and force profiles from the real data. 
We investigated modeling the tire as a perfect cylinder, a perfect torus, 
or a convex triangular mesh. 
Although a perfect torus should ideally match the wheel geometry on the real robot, 
we found the mesh approximation worked better on the robot. 
We suspect the approximation induced some noise in the ground--wheel contact interaction, 
which helped make the policies more robust in the real world.

\section{Actuators Modeling}\label{sec_actuators}
To better match the physical actuators, 
the simulation incorporates realistic torque--speed characteristics. 
Joint torque limits are computed dynamically from the motors' electromechanical constraints, 
including maximum current, bus voltage, and torque constant, 
rather than being imposed as fixed saturation bounds. 
The torque constants were experimentally identified using a custom dynamometer test rig. 
To identify the nominal PD gains in~\tref{table_act}, 
the bicycle base was fixed and sinusoidal position and velocity commands were applied to each joint. 
This approach evaluates tracking performance over continuously varying trajectories rather than static setpoints. 
The gains were tuned by analyzing the phase lag and amplitude attenuation of the measured joint response 
across a range of excitation frequencies, 
yielding the desired closed-loop bandwidth. 
Additionally, the repeated direction reversals of the sinusoidal trajectories expose hardware nonlinearities, 
including gear backlash, friction, and motor saturation.

\section{State Estimation}\label{sec_state_est}
The state estimate consists of the 6-DoF floating-base state and the six serial-joint states. 
The quantities used to construct the observations in~\tref{table_obs_act}(A,B) 
include the joint positions and velocities, 
the bike base global position (for tasks that require it), 
the base orientation parameterized as a ($3\times3$) rotation matrix (used to compute projected gravity), 
and the base angular velocity. 
Joint positions and velocities ($q_s$, $q_w$, $q_h$, $q_n$) are measured directly from encoders, 
while the angular velocity is obtained from the onboard~\gls{imu} gyroscope. 
To estimate the floating-base state, we fuse~\gls{imu} and~\gls{mocap} measurements, 
the latter providing accurate global position during laboratory experiments 
though not required for all tasks; 
the robot also supports a fully onboard estimator for outdoor operation~\cite{RAIInstitute2026UMV}.

State estimation uses the GTSAM~\cite{gtsam} factor-graph optimization framework 
following a bipartite architecture similar to~\cite{goldfain2019autorally}. 
\gls{imu} angular rates and linear accelerations are integrated at \unit[1]{kHz} 
for a low-latency estimate, 
while motion capture measurements are incorporated at \unit[30]{Hz} to correct drift. 
The factor graph estimates position, orientation, linear velocity, and inertial sensor biases. 
Including motion-capture orientation measurements improves estimator robustness during highly dynamic maneuvers, 
where significant~\gls{imu} noise or saturation may occur. 
The estimator runs at \unit[1]{kHz} and is optimized at \unit[30]{Hz} 
using the latest motion capture measurements and pre-integrated~\gls{imu} data~\cite{imu_preintegration}.

The waypoint following task requires a robot-centric heightmap observation, 
extracted from a larger 2.5D grid map represented in a global frame. 
Because the current platform lacks onboard exteroceptive sensors, 
this global grid map is generated offline before each experiment from the known terrain configuration. 
During operation, the estimated robot pose localizes the robot within the global grid map 
to extract the local heightmap, 
which ultimately relies on the motion capture system through the state-estimation pipeline above.

\section{Orchestrator Details}\label{sec_orch}
Several common approaches could coordinate these skills, 
namely~\glspl{fsm}, \glspl{bt}, 
and funnel-based sequential composition methods~\cite{burridge1999sequential}, 
which operate at different levels of abstraction: 
\glspl{fsm} and \glspl{bt} primarily define logical relationships between behaviors, 
while sequential composition defines dynamical relationships between regions of the state space. 
We therefore designed the orchestrator based on sequential composition~\cite{burridge1999sequential}. 
Each policy is associated with a funnel or region of attraction, 
and transitions are permitted only when the terminal set of one policy lies 
within the admissible initial set of the next. 
Thus one policy prepares another if it can drive the robot into a 
region of state space satisfying the entry conditions of the subsequent policy, 
and a transition becomes feasible whenever the policy domains overlap sufficiently 
for both to remain valid during the switch. 
The orchestrator transitions are designed to identify these overlap regions 
and trigger policy switches only when the required preconditions are satisfied.

\begin{figure*}
\centering
\includegraphics[width=0.99\textwidth]{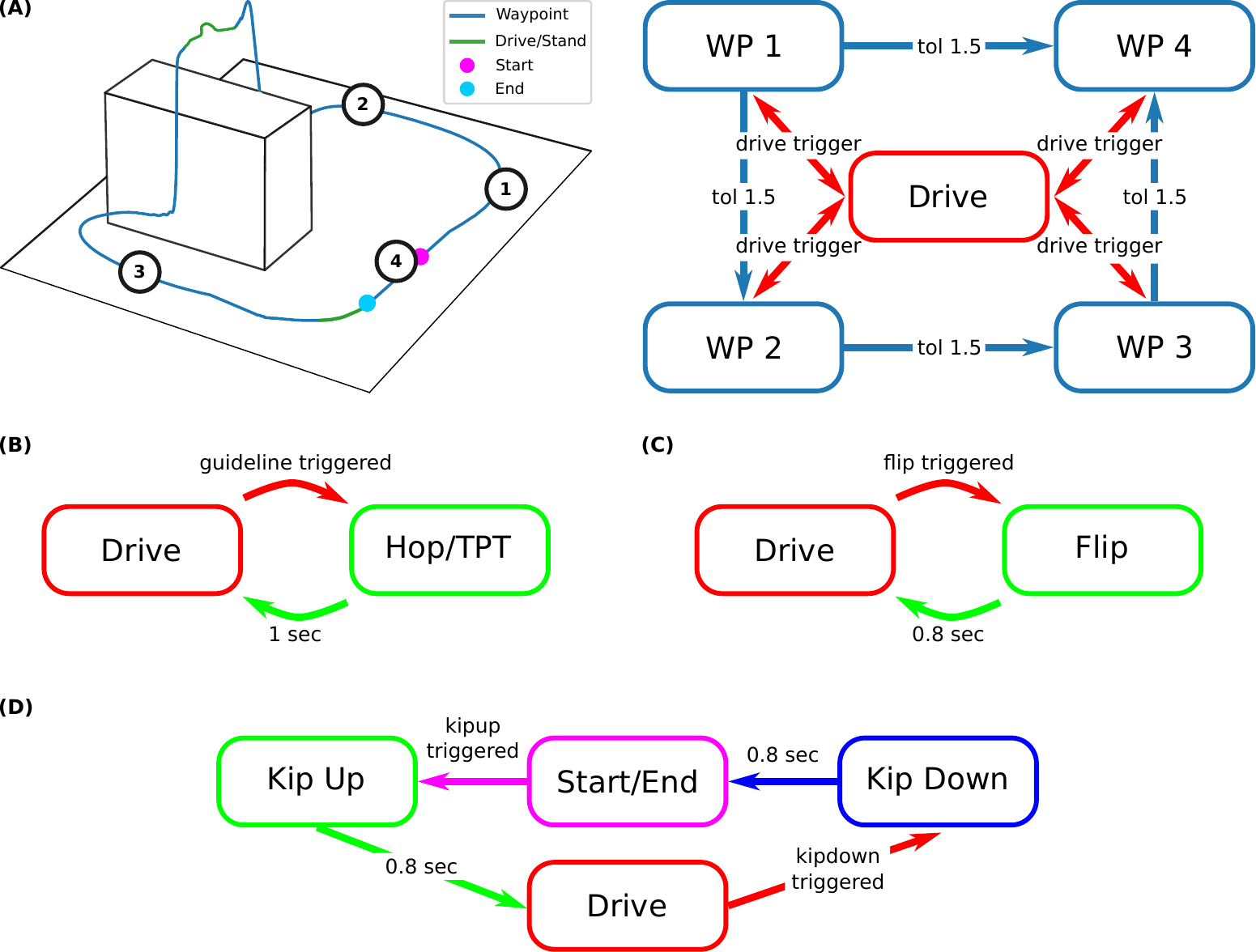}
\caption{\textbf{Additional Orchestrator Diagrams.} 
(\textbf{A})~Jump and track stand (\tref{tab_robot_stunts_taxonomy}(\textbf{16})
and \fref{fig_all_jumps}(\textbf{E}))
where the jump policy is the main running policy 
and a joystick button can transition to a drive policy for emergency user takeover
which allowed the robot to stand on the table. 
(\textbf{B},\textbf{C})~Stunt-and-drive stunts 
(\tref{tab_robot_stunts_taxonomy}(\textbf{17},\textbf{19},\textbf{20})
and 
\fref{fig_other_stunts}(\textbf{D},\textbf{E},\textbf{F})), 
where driving is plugged in as an emergency controller. 
(\textbf{D})~Kip-and-drive 
stunt (\tref{tab_robot_stunts_taxonomy}(\textbf{18})
and \fref{fig_other_stunts}(\textbf{C})), 
where kip-up, user-driven driving, and kip-down policies are sequenced and can repeat.}
\label{fig_orch_odnt_supp}
\end{figure*}

Here we give more examples of using the orchestrator. 
\fref{fig_orch_odnt_supp} details the orchestrator structure for four different stunts. 
\fref{fig_orch_odnt_supp}(A) shows the orchestrator diagram of a typical 
repetitive~\unit[75]{cm} and~\unit[1]{m} jumps
where the main policy that is running is the jump policy. 
However, the orchestrator is structured so that if a joystick button is triggered, 
the jump policy transitions to a drive policy as an emergency stunt where a user can take control over the robot. 
An example of using this is the track stand on the table in~\fref{fig_all_jumps}(E)
which is highlighted in the 3D plot of the robot trajectory in~\fref{fig_orch_odnt_supp}(A).

In similar fashions, 
we can plug in the drive 
not just as an emergency controller to any of the other policies,
but also as a way to plug in any of the policies trained with guided tracking or motion imitation. 
For example,~\fref{fig_orch_odnt_supp}(B,C) show the orchestrator diagrams 
that are used for the stunt-and-drive stunts~\tref{tab_robot_stunts_taxonomy}(17,19,20).

\fref{fig_orch_odnt_supp}(D) shows the orchestrator diagram of the
kip and drive stunt~\fref{fig_other_stunts}(C). 
In this stunt, the robot always starts in an upright position, and once a kip up button is triggered, 
the orchestrator switches to the kip up policy then to the drive policy where the robot is driven by a user. 
The user can trigger a kip down policy that puts the robot back into the upright configuration. 
This can happen repetitively.

\clearpage

\begin{table}[t]
\centering
\caption{\bf Observations \& Actions.}
\resizebox{1\linewidth}{!}{
\begin{tabular}{l ccccccc}
\toprule 
\textbf{\shortstack[l]{Term\\~}} & \textbf{\shortstack{Dim.\\~}} & \textbf{\shortstack{Waypoint\\Following}} & \textbf{\shortstack{Pose\\Reaching}} & \textbf{\shortstack{SE2 Twist\\Tracking}} & \textbf{\shortstack{Guided\\Tracking}} & \textbf{\shortstack{Motion\\Imitation}} & \textbf{\shortstack{Noise$^1$\\~}} \\
\midrule
\ourrow \multicolumn{8}{l}{\textbf{(A) Actor (Policy) Observations: Common Observations}} \\
Bike Angular Velocity~($\omega$)         & 3 & \Checkmark & \Checkmark & \Checkmark & \Checkmark & \Checkmark & $\mathcal{U}(-0.1, 0.1)$ \\
Projected Gravity~($g$)                  & 3 & \Checkmark & \Checkmark & \Checkmark & \Checkmark & \Checkmark & $\mathcal{U}(-0.015, 0.015)$ \\
Steering Joint Position~($q_s$)          & 1 & \Checkmark & \Checkmark & \Checkmark & \Checkmark & \Checkmark & $\mathcal{N}(0, 0.001)$ \\
Steering Joint Velocity~($\dot{q}_s$)    & 1 & \Checkmark & \Checkmark & \Checkmark & \Checkmark & \Checkmark & $\mathcal{N}(0, 0.1)$ \\
Rear Wheel Joint Velocity~($\dot{q}_w$)  & 1 & \Checkmark & \Checkmark & \Checkmark & \Checkmark & \Checkmark & $\mathcal{N}(0, 0.5)$ \\
Head Joint Position~($q_h$)              & 1 & \Checkmark & \Checkmark & \Checkmark & \Checkmark & \Checkmark & $\mathcal{N}(0, 0.001)$ \\
Head Joint Velocity~($\dot{q}_h$)        & 1 & \Checkmark & \Checkmark & \Checkmark & \Checkmark & \Checkmark & $\mathcal{N}(0, 0.22)$ \\
Neck Joint Position~($q_n$)              & 1 & \Checkmark & \Checkmark & \Checkmark & \Checkmark & \Checkmark & $\mathcal{N}(0, 0.001)$ \\
Neck Joint Velocity~($\dot{q}_n$)        & 1 & \Checkmark & \Checkmark & \Checkmark & \Checkmark & \Checkmark & $\mathcal{N}(0, 0.22)$ \\
Last Actions~($a_{t-1}$)                 & 4 & \Checkmark & \Checkmark & \Checkmark & \Checkmark & \Checkmark & -- \\
\midrule
\ourrow \multicolumn{8}{l}{\textbf{(B) Actor (Policy) Observations: Policy-Specific Observations}} \\
2D Goal Position Command ($x_{\text{goal}}$)     & 2   & \Checkmark   & \XSolidBrush & \XSolidBrush & \XSolidBrush & \XSolidBrush & -- \\
SE2 Velocity Command ($v_c$)                     & 2   & \XSolidBrush & \XSolidBrush & \Checkmark   & \Checkmark   & \XSolidBrush & -- \\
Phase Clock Command  ($\phi$)                    & 1   & \XSolidBrush & \XSolidBrush & \XSolidBrush & \XSolidBrush & \Checkmark   & -- \\
Stunt Trigger Command ($c_{\text{stunt}}$)       & 1   & \XSolidBrush & \Checkmark   & \XSolidBrush & \Checkmark   & \XSolidBrush & -- \\
Bike Height   ($z_{\text{bike}}$)                & 1   & \Checkmark   & \XSolidBrush & \XSolidBrush & \Checkmark   & \XSolidBrush & $\mathcal{N}(-0.1, 0.1)$ \\
Bike Global Position ($x_{\text{bike}}$)         & 2   & \XSolidBrush & \XSolidBrush & \XSolidBrush & \Checkmark   & \XSolidBrush & $\mathcal{N}(0, 0.1)$ \\
Heightmap ($H_1$)$^2$                              & 112 & \Checkmark   & \XSolidBrush & \XSolidBrush & \XSolidBrush & \XSolidBrush & Bias: $\mathcal{U}(-0.05, 0.05)$\\
~                                                &~    &~             &~             &~             &~             &~             & Drift: $\mathcal{U}(-0.1, 0.1)$\\
\midrule
\ourrow \multicolumn{8}{l}{\textbf{(C) Additional Critic (Privileged) Observations}} \\
Bike Linear Velocity ($v_{\text{bike}}$) & 3 & \Checkmark & \Checkmark   & \Checkmark   & \Checkmark   & \Checkmark   & -- \\
Terrain Difficulty Level                 & 1 & \Checkmark & \XSolidBrush & \XSolidBrush & \XSolidBrush & \XSolidBrush & -- \\
Heightmap ($H_2$)$^3$                              & 112 & \Checkmark   & \XSolidBrush & \XSolidBrush & \XSolidBrush & --\\
\midrule
\ourrow \multicolumn{8}{l}{\textbf{(D) Actions}} \\
Steering Joint Position Setpoint~($a_s$)         & 1 & \Checkmark & \Checkmark & \Checkmark & \Checkmark & \Checkmark & -- \\
Rear Wheel Joint Velocity Setpoint~($a_w$)       & 1 & \Checkmark & \Checkmark & \Checkmark & \Checkmark & \Checkmark & -- \\
Head Joint Position Setpoint~($a_h$)             & 1 & \Checkmark & \Checkmark & \Checkmark & \Checkmark & \Checkmark & -- \\
Neck Joint Position Setpoint~($a_n$)             & 1 & \Checkmark & \Checkmark & \Checkmark & \Checkmark & \Checkmark & -- \\
\bottomrule
\multicolumn{8}{l}{$^1$ The terms~$\mathcal{U}$ and~$\mathcal{N}$ denote the Uniform and Gaussian distributions, respectively.} \\
\multicolumn{8}{l}{$^2$ The Heightmap~($H_1$) is $1.5m \times0.6m$ with a resolution of $0.1m$.} \\
\multicolumn{8}{l}{$^3$ The Heightmap~($H_2$) is $3.0m \times1.2m$ with a resolution of $0.2m$.} \\
\end{tabular}
}
\label{table_obs_act}
\end{table}

\begin{table}[t]
\centering
\caption{\bf Actions Gains.}
\mbox{
\begin{tabular}{l ccccc}
\toprule 
\textbf{Term} & \textbf{Waypoint Following} & \textbf{Pose Reaching} & \textbf{SE2 Twist Tracking} & \textbf{Guided Tracking} & \textbf{Motion Imitation} \\
\midrule
\ourrow \multicolumn{6}{l}{\textbf{(A) Action Scales}} \\
Steering Joint   & 1   & 1   & 1   & 1   & 1    \\
Rear Wheel Joint & 15  & 15  & 15  & 15  & 15   \\
Head Joint       & 0.1 & 1.5 & 0.1 & 0.1 & 0.1  \\
Neck Joint       & 0.1 & 1.5 & 0.1 & 0.1 & 0.1  \\
\midrule
\ourrow \multicolumn{6}{l}{\textbf{(B) Action Stiffness}} \\
Steering Joint   & 1.4 & 1.4 & 1.4 & 1.4 & 1.4 \\
Rear Wheel Joint & 0   & 0   & 0   & 0   & 0   \\
Head Joint       & 140 & 140 & 120 & 200 & 200 \\
Neck Joint       & 140 & 140 & 120 & 200 & 200 \\
\midrule
\ourrow \multicolumn{6}{l}{\textbf{(C) Action Damping}} \\
Steering Joint   & 0.15 & 0.15 & 0.15 & 0.15 & 0.15 \\
Rear Wheel Joint & 0.6  & 0.6  & 0.6  & 0.6  & 0.6  \\
Head Joint       & 10   & 10   & 5    & 5    & 5    \\
Neck Joint       & 10   & 10   & 5    & 5    & 5    \\
\bottomrule
\end{tabular}
}
\label{table_act}
\end{table}

\begin{table*}[t]
\centering
\caption{\bf Waypoint Following Jump Rewards.}
\resizebox{1\linewidth}{!}{
\begin{tabular}{l c c l }
\toprule 
\textbf{Term} & \textbf{Expression} & \textbf{Weight} & \textbf{Description} \\
\midrule 
\ourrow \multicolumn{4}{l}{\textbf{(A) Task Rewards}} \\ 
Distance to goal tracking       & $\mathrm{exp}(- \left\| x^\mathrm{bike}_\mathrm{goal} \right\| )$  & $6$   & Track distance to target goal position.\\
Angle to goal tracking$^1$      & $\cos(\theta_{\text{goal}}^{\text{bike}})$          & $1.5$ & Move towards the direction of the goal.\\
\midrule 
\ourrow \multicolumn{4}{l}{\textbf{(B) Style Rewards}} \\ 
Stay Alive                      & 1                                              & $10$   & Reward the robot for being alive.\\
Non-flat orientation penalty    & $|g_x|$                                        & $-1.0$ & Penalize lateral body roll.\\
High linear velocity penalty    & $ \max\{ (v_{\mathrm{bike}} - 2.5)^2 , 0\}$    & $-3.0$ & Penalize forward velocities exceeding 2.5 m/s.\\
Angular velocity penalty        & $ |\omega_z | $                                & $-0.1$ & Penalize excessive yaw rates.\\
Joint $q_h$ near zero reward    & $|q_h - 0| $                                   & $1$    & Track zero head joint position.\\
Joint $q_n$ near zero reward    & $|q_n - 0| $                                   & $1$    & Track zero neck joint position.\\
\midrule
\ourrow \multicolumn{4}{l}{\textbf{(C) Regularization Rewards}} \\ 
Joint $q_s$ action rate$^2$     & $|\Delta a_s | $                              & $-1\times10^{-4}$   & Penalize large changes in the processed action.\\
Joint $q_w$ action rate$^3$     & $|\Delta a_w | $                              & $-1\times10^{-5}$   & Penalize large changes in the processed action.\\
Joint $q_h$ action rate$^4$     & $|\Delta a_h | $                              & $-1\times10^{-4}$   & Penalize large changes in the processed action.\\
Joints $q_n$ action rate$^4$    & $|\Delta a_n| $                               & $-1\times10^{-4}$   & Penalize large changes in the processed action.\\
Joint torques$^4$               & $\Sigma_j|\tau_j| $                           & $-1\times10^{-4}$   & Penalize large joint torques.\\
Joint $q_s$ velocity            & $|\dot{q}_s | $                               & $-1\times10^{-3}$   & Penalize high steering fork velocity.\\
Joint $q_h$, $q_n$ velocity     & $|\dot{q}_h| + |\dot{\mu}| $                  & $-1\times10^{-2}$   & Penalize high head and neck joint velocities.\\
Joint $q_h$ upper limit   & $|q_h - q_{h, \max}|/0.05 $         & $-2$   & Penalize exceeding the head joint maximum limit.\\
Joint $q_h$ lower limit   & $|q_h - q_{h, \min} |/0.03  $       & $-2$   & Penalize exceeding the head joint minimum limit.\\
Joint $q_n$ upper limit   & $|q_n - q_{n, \max}|/0.05 $         & $-2$   & Penalize exceeding the neck joint maximum limit.\\
Joint $q_n$ lower limit   & $|q_n - q_{n, \min} |/0.03  $       & $-2$   & Penalize exceeding the neck joint minimum limit.\\
\bottomrule
\multicolumn{4}{l}{\small$^1$The term $\cos(\theta_{\text{goal}}^{\text{bike}})$ is defined as $(v_\mathrm{bike} \cdot  x^\mathrm{bike}_\mathrm{goal})     /     (\|v_\mathrm{bike}\| \, \| x^\mathrm{bike}_\mathrm{goal}\|)$.}\\
\multicolumn{4}{l}{\small$^2$The action rate $|\Delta a_j|$ is defined as the change in the processed action for the $j$th joint $a_j - a_{j, t-1}$.}\\
\multicolumn{4}{l}{\small$^3$The weight of this reward is initialized with $-1\times10^{-7}$ and then modified to $-1\times10^{-5}$ as a curriculum after 5000 iterations during training.}\\
\multicolumn{4}{l}{\small$^4$These rewards are applied later as a curriculum after 5000 iterations during training.}\\
\end{tabular}
}
\label{table_jump_reward_functions} 
\end{table*}

\begin{table*}[t]
\centering
\caption{\bf Pose Reaching Kip Up \& Down Rewards.}
\resizebox{1\linewidth}{!}{
\begin{tabular}{l c c l }
\toprule 
\textbf{Term} & \textbf{Expression} & \textbf{Weight} & \textbf{Description} \\
\midrule 
\ourrow \multicolumn{4}{l}{\textbf{(A) Task Rewards: Upside Down (reward the robot to be and stay upside down)}} \\ 
Upside Down Contacts Reward    & $\mathbb{I}(F_{\text{head}} > \epsilon_1) \cdot \mathbb{I}(F_{\text{wheels}} \le \epsilon_2)$  & $0.1$   & Encourage head and discourage wheel contact with the ground.\\
Upside Down Orientation Reward & $|g_z|$                                                                                        & $0.1$   & Encourage upside-down orientation.\\
\midrule 
\ourrow \multicolumn{4}{l}{\textbf{(B) Task Rewards: Upright (reward the robot to be and stay upright) }} \\ 
Upright Contacts Reward         & $ \mathbb{I}(F_{\text{head}} \le \epsilon_3) \cdot \mathbb{I}(F_{\text{wheels}} > \epsilon_4)$  & $5$      & Encourage wheel and discourage head contact with the ground.\\
Upright Bike Orientation Reward & $-|g_z|$                                                                                        & $1$      & Encourage upright bike orientation.\\
Upright Head Orientation Reward & $-|g_{\text{head}, z}|$                                                                         & $0.25$   & Encourage upright head orientation.\\
Upright Contacts Penalty        & $ \mathbb{I}(F_{\text{head}} > \epsilon_5)$                                                     & $-3$     & Penalize high head contact with the ground.\\
Base Linear Velocity Tracking   & $\mathrm{exp}(-| v^\mathrm{x}_\mathrm{base} |) $                                                & $2$      & Track zero forward linear velocity of the robot base.\\
Heading Tracking                & $\mathrm{exp}(-| \omega^\mathrm{z}_\mathrm{base} |) $                                           & $2$      & Track zero angular (yaw) velocity of the robot base. \\
\midrule 
\ourrow \multicolumn{4}{l}{\textbf{(C) Style Rewards }} \\ 
Joint $q_h$ near zero reward    & $|q_h - 0| $                                   & $1$    & Track zero head joint position.\\
Joint $q_n$ near zero reward    & $|q_n - 0| $                                   & $1$    & Track zero neck joint position.\\
\midrule 
\ourrow \multicolumn{4}{l}{\textbf{(D) Regularization Rewards (Kip Up and Kip Down)}} \\ 
Joint $q_s$ action rate$^1$                & $|\Delta a_s| $                              & $-1\times10^{-4}$    & Penalize large changes in the processed action.\\
Joint $q_w$ action rate$^1$                & $|\Delta a_w| $                              & $-1\times10^{-8}$    & Penalize large changes in the processed action.\\
Joint $q_h$ action rate$^1$                & $|\Delta a_h| $                              & $-1\times10^{-4}$    & Penalize large changes in the processed action.\\
Joints $q_n$ action rate$^1$               & $|\Delta a_n| $                              & $-1\times10^{-4}$    & Penalize large changes in the processed action.\\
Joint torques$^1$                          & $\Sigma_j|\tau_j| $                           & $-1\times10^{-4}$   & Penalize large joint torques.\\
Joint velocities$^1$                       & $\Sigma_j|\dot{q}_i | $                       & $-1\times10^{-4}$   & Penalize large joint velocities.\\
Joint $q_h$ upper limit                & $|q_h - q_{h, \max}|/0.05 $                   & $-1$                 & Penalize exceeding the upper limit.\\
Joint $q_h$ lower limit                & $|q_h - q_{h, \min} |/0.03  $                 & $-1$                 & Penalize exceeding the lower limit.\\
Joint $q_n$ upper limit                & $|q_n - q_{n, \max}|/0.05 $                   & $-1$                 & Penalize exceeding the upper limit.\\
Joint $q_n$ lower limit                & $|q_n - q_{n, \min} |/0.03  $                 & $-1$                 & Penalize exceeding the lower limit.\\
Joint $q_r$ upper limit                & $|q_r - q_{r, \max}|/ 0.1$                   & $-1.5$                & Penalize exceeding the upper limit.\\
Joint $q_r$ lower limit                & $|q_r - q_{r, \min} |/0.1  $                 & $-1.5$                & Penalize exceeding the lower limit.\\
Joint $q_l$ upper limit                & $|q_l - q_{l, \max}|/ 0.1$                   & $-1.5$                & Penalize exceeding the upper limit.\\
Joint $q_l$ lower limit                & $|q_l - q_{l, \min}|/ 0.1  $                 & $-1.5$                & Penalize exceeding the lower limit.\\
\bottomrule
\multicolumn{4}{l}{\small $^1$The weights of these rewards are initialized with $-1\times10^{-12}$ and then modified to weights above as a curriculum after 2000 iterations during training.}\\
\end{tabular}
}
\label{table_kip_reward_functions} 
\end{table*}

\begin{table*}[t]
\centering
\caption{\bf SE2 Twist Tracking Drive Rewards.}
\resizebox{1\linewidth}{!}{
\begin{tabular}{l c c l }
\toprule 
\textbf{Term} & \textbf{Expression} & \textbf{Weight} & \textbf{Description} \\
\midrule 
\ourrow \multicolumn{4}{l}{\textbf{(A) Task Rewards}} \\ 
Bike linear velocity tracking          & $\mathrm{exp}(-| v^\mathrm{x}_\mathrm{bike} -v_\mathrm{c}^\mathrm{x} |) $ & $3$ & Track the forward linear velocity of the robot base. \\
Bike heading tracking$^1$      & $\mathrm{exp}(-| \omega^\mathrm{z}_\mathrm{bike} - \omega_\mathrm{c}^\mathrm{z} |) $ & $1.5$ & Track the angular (yaw) velocity of the robot base. \\
Stay Alive                             & 1                                                                                    & $2$   & Constant reward for continuous forward driving.\\
\midrule
\ourrow \multicolumn{4}{l}{\textbf{(C) Regularization Rewards}} \\ 
Joint $q_s$ action rate                & $|\Delta a_s | $                              & $-2\times10^{-4}$          & Penalize large changes in the processed action.\\
Joint $q_w$ action rate$^*$            & $|\Delta a_w | $                              & $-2\times10^{-5}$          & Penalize large changes in the processed action.\\
Joint $q_h$ action rate               & $|\Delta a_h | $                              & $-1\times10^{-3}$   & Penalize large changes in the processed action.\\
Joints $q_n$ action rate             & $|\Delta a_n| $                               & $-1\times10^{-3}$   & Penalize large changes in the processed action.\\
Joint $q_h$, $q_n$ torques            & $|\tau_h| + |\tau_n| $                  & $-1\times10^{-3}$   & Penalize high head and neck joint torques.\\
Joint $q_s$ velocity                   & $|\dot{q}_s | $                               & $-1\times10^{-3}$   & Penalize high steering fork velocity.\\
Joint $q_h$, $q_n$ velocity            & $|\dot{q}_h| + |\dot{\mu}| $                  & $-1\times10^{-1}$   & Penalize high head and neck joint velocities.\\
Joint $q_h$ limits                & $\mathbb{I}(q_h \notin [q_{\text{min, h}}, \ q_{\text{max, h}}])$                 & $-2$                & Penalize head joint out of limits.\\
Joint $q_n$ limits                & $\mathbb{I}(q_n \notin [q_{\text{min, n}}, \ q_{\text{max, n}}])$                 & $-2$                & Penalize neck joint out of limits.\\
\bottomrule
\multicolumn{4}{l}{\small$^1$These rewards are applied later as a curriculum after 2000 iterations during training.}\\
\end{tabular}
}
\label{table_drive_reward_functions} 
\end{table*}

\begin{table*}[t]
\centering
\caption{\bf Guided Tracking Rewards.}
\resizebox{1\linewidth}{!}{
\begin{tabular}{l c c l }
\toprule 
\textbf{Term} & \textbf{Expression} & \textbf{Weight} & \textbf{Description} \\
\midrule 
\ourrow \multicolumn{4}{l}{\textbf{(A) Task Rewards}} \\ 
Guideline tracking              & $\|x^{\text{stunt}}_{\text{t}} - p_{\text{goal}}\|_2 - \|x^{\text{stunt}}_{t-1} - p_{\text{goal}}\|_2$ & $200$   & Track the guideline reference.\\
Keyframe Orientation tracking   & $\exp\!\left(2\,\arccos\!\left( \left| q_t \cdot q_i \right| \right)\right)$ or $-(\theta^{\text{diff}}_t - \theta^{\text{diff}}_{t-1})$  & $150$ & Track orientation keyframes.\\
\midrule 
\ourrow \multicolumn{4}{l}{\textbf{(B) Style Rewards}} \\ 
Non-flat orientation penalty           & $|g_x|$                                       & $-1.0$  & Penalize lateral body roll.\\
High linear velocity penalty           & $ \max\{ (v_{\mathrm{bike}} - 2.5)^2 , 0\}$   & $-3.0$  & Penalize forward velocities exceeding 2.5 m/s.\\
Angular velocity penalty               & $ |\omega^\mathrm{z}_\mathrm{bike} | $        & $-0.1$  & Penalize excessive yaw rates.\\
Joint $q_h$ near zero reward           & $|q_h - 0| $                                  & $1$     & Track zero head joint position.\\
Joint $q_n$ near zero reward           & $|q_n - 0| $                                  & $1$     & Track zero neck joint position.\\
\midrule
\ourrow \multicolumn{4}{l}{\textbf{(C) Regularization Rewards}} \\ 
Joint $q_s$ velocity                  & $|\dot{q}_s | $                                  & $-1\times10^{-3}$    & Penalize high steering fork velocity.\\
Joint power                           & $\Sigma_j|\tau_j\cdot\dot{q}_j| $                & $-1\times10^{-7}$    & Penalize high mechanical power consumption.\\
Joint $q_s$ action rate               & $|\Delta a_s | $                                 & $-1\times10^{-4}$    & Penalize large changes in the processed action.\\
Joint $q_w$ action rate$^1$           & $|\Delta a_w | $                                 & $-1\times10^{-5}$    & Penalize large changes in the processed action.\\
Joint $q_h$ action rate$^1$           & $|\Delta a_h | $                                 & $-1\times10^{-5}$    & Penalize large changes in the processed action.\\
Joints $q_n$ action rate$^1$          & $|\Delta a_n| $                                  & $-1\times10^{-5}$    & Penalize large changes in the processed action.\\
Joint torques$^1$                     & $\Sigma_j|\tau_j| $                              & $-1\times10^{-5}$    & Penalize high joint torques.\\
Joint $q_h$ upper limit$^1$           & $|q_h - q_{h, \max}|/0.05 $                      & $-10$                & Penalize exceeding the upper limit.\\
Joint $q_h$ lower limit$^1$           & $|q_h - q_{h, \min} |/0.03  $                    & $-10$                & Penalize exceeding the lower limit.\\
Joint $q_n$ upper limit$^1$           & $|q_n - q_{n, \max}|/0.05 $                      & $-10$                & Penalize exceeding the upper limit.\\
Joint $q_n$ lower limit$^1$           & $|q_n - q_{n, \min} |/0.03  $                    & $-10$                & Penalize exceeding the lower limit.\\
Joint $q_w$ Velocity limit$^1$        & $\mathbb{I}(\dot{q}_w > 77\text{rad/s})$         & $-1\times10^{-5}$    & Penalize rear wheel velocity exceeding 77 rad/s.\\
High Contact Forces Penalty$^1$       & $ \mathbb{I}(F_{\text{wheels}} > 350\text{N})$   & $-1\times10^{-5}$    & Penalize high contact forces exceeding a certain threshold.\\
\bottomrule
\multicolumn{4}{l}{\small $^1$These rewards are activated as a curriculum after the robot first successfully learns and completes the full guideline reference.}\\
\end{tabular}
}
\label{table_guideline_reward_functions} 
\end{table*}

\begin{table*}[t]
\centering
\caption{\bf Motion Imitation Rewards.}
\resizebox{1\linewidth}{!}{
\begin{tabular}{l c c l }
\toprule 
\textbf{Term} & \textbf{Expression} & \textbf{Weight} & \textbf{Description} \\
\midrule 
\ourrow \multicolumn{4}{l}{\textbf{(A) Task Rewards: Imitation}} \\ 
Base Position Reference Tracking & $\exp(\| p_{\text{base}} - p_{\text{base}}^{\text{ref}} \|^2)$ & 4.0 & Track the robot base position reference.\\
Base Orientation Reference Tracking & $\exp(-\| d_\text{angle}(q_{\text{base}},q_{\text{base}}^{\text{ref}}) \|^2)$ & 20.0 & Track the robot base orientation reference.\\
Joint Positions Reference Tracking & $\exp(-\| q_{\text{joint}} - q_{\text{joint}}^{\text{ref}} \|^2)$ & 1.0 & Track the robot joint position references.\\
\midrule 
\ourrow \multicolumn{4}{l}{\textbf{(B) Task Rewards: Post-Imitation}} \\ 
Default Joint Positions & $\exp(-\alpha_{\text{pd}} \| q_{\text{body joint}} \|^2)$ & 1.0 & Track default joint positions.\\
Jitter Penalty & $\sum \dot{q}_{\text{joint}}^2 + \sum |\dot{q}_{\text{joint}}|$ & -1e-4 & Penalize high joint velocities to reduce jitter.\\
Velocity Penalty & $\| v_{\text{base}} \|^2$ & -3.0 & Penalize high velocities.\\
\midrule
\ourrow \multicolumn{4}{l}{\textbf{(C) Regularization Rewards}} \\ 
Joint $q_s$ action rate        & $|\Delta a_s | $                                                      & $-1\times10^{-4}$   & Penalize large changes in the processed action.\\
Joint $q_w$ action rate$^*$    & $|\Delta a_w | $                                                      & $-1\times10^{-5}$   & Penalize large changes in the processed action.\\
Joint $q_h$ action rate$^1$    & $|\Delta a_h | $                                                      & $-1\times10^{-4}$   & Penalize large changes in the processed action.\\
Joints $q_n$ action rate$^1$   & $|\Delta a_n| $                                                       & $-1\times10^{-4}$   & Penalize large changes in the processed action.\\
Joint torques$^1$              & $\Sigma_j|\tau_j| $                                                   & $-1\times10^{-5}$   & Penalize high joint torques.\\
Joint $q_s$ velocity           & $|\dot{q}_s | $                                                       & $-1\times10^{-3}$   & Penalize high steering fork velocity.\\
Joint $q_h$, $q_n$ velocity    & $|\dot{q}_h| + |\dot{\mu}| $                                          & $-1\times10^{-2}$   & Penalize high head and neck joint velocities.\\
High Contact Forces Penalty    & $ \mathbb{I}(F_{\text{wheels}} > 350\text{N})$                        & $-1\times10^{-5}$   & Penalize high contact forces exceeding a certain threshold.\\
Joint $q_h$ limits             & $\mathbb{I}(q_h \notin [q_{\text{min, h}}, \ q_{\text{max, h}}])$     & $-1$                & Penalize head joint out of limits.\\
Joint $q_n$ limits             & $\mathbb{I}(q_n \notin [q_{\text{min, n}}, \ q_{\text{max, n}}])$     & $-1$                & Penalize neck joint out of limits.\\
\bottomrule
\multicolumn{4}{l}{\small $^1$These rewards are activated as a curriculum after the robot first successfully learns and completes the full guideline reference.}\\
\end{tabular}
}
\label{table_flip_reward_functions} 
\end{table*}

\begin{table*}[t]
\centering
\caption{\bf Domain Randomization}
\label{table_domain_randomization_complete}
\resizebox{\linewidth}{!}{
\begin{tabular}{l c ccccc}
\toprule
\textbf{\shortstack[l]{Term\\~}} & \textbf{\shortstack[l]{Dim.\\~}} & \textbf{\shortstack{Waypoint\\Following}} & \textbf{\shortstack{Pose\\Reaching}} & \textbf{\shortstack{SE2 Twist\\Tracking}} & \textbf{\shortstack{Guided\\Tracking}} & \textbf{\shortstack{Motion\\Imitation}} \\
\midrule
\ourrow\multicolumn{7}{l}{\textbf{(A) Actuation \& Rigid Body Dynamics}} \\
Friction Coefficient & $-$ & $\mathcal{U}[0.6, 1.6]$ & $\mathcal{U}[0.6, 1.6]$ & $\mathcal{U}[0.7, 1.6]$ & $\mathcal{U}[0.6, 1.6]$ & $\mathcal{U}[0.6, 1.6]$ \\
Added Mass Scale & ratio & $\mathcal{U}[0.9, 1.1]$ & $\mathcal{U}[0.9, 1.1]$ & $\mathcal{U}[0.9, 1.1]$ & $\mathcal{U}[0.9, 1.1]$ & $\mathcal{U}[0.9, 1.1]$ \\
Added Center of Mass & m & $\mathcal{U}[-0.02, 0.02]$ & $-$ & $-$ & $\mathcal{U}[-0.02, 0.02]$ & $\mathcal{U}[-0.02, 0.02]$ \\
Motor Strength Scale & ratio & $\mathcal{U}[0.85, 1.05]$ & $\mathcal{U}[0.85, 1.05]$ & $-$ & $\mathcal{U}[0.85, 1.05]$ & $\mathcal{U}[0.85, 1.05]$ \\
Actuation Delay & steps & $\mathcal{U}[0, 1]$ & $\mathcal{U}[0, 1]$ & $-$ & $\mathcal{U}[0, 1]$ & $\mathcal{U}[0, 1]$ \\
Joint PD Gains Scale & ratio & $-$ & $\mathcal{U}[0.85, 1.15]$ & $-$ & $-$ & $\mathcal{U}[0.85, 1.15]$ \\
Joint Friction Coefficient & $-$ & $\mathcal{U}[0.0, 0.04]$ & $\mathcal{U}[0.0, 0.04]$ & $-$ & $-$ & $\mathcal{U}[0.0, 0.04]$ \\
\midrule
\ourrow\multicolumn{7}{l}{\textbf{(B) Initial Base Pose \& Twist State Randomization$^1$}} \\
Bike Base Position $X$            & m   & $\mathcal{U}[-0.1, 0.1]$   & $-$              & $\mathcal{U}[-0.1, 0.1]$   & $\mathcal{U}[-0.03, 0.03]$ & $\mathcal{U}[-0.02, 0.02]$ + ref.\\
Bike Base Position $Y$            & m   & $\mathcal{U}[-0.1, 0.1]$   & $-$              & $\mathcal{U}[-0.1, 0.1]$   & $\mathcal{U}[-0.03, 0.03]$ & $\mathcal{U}[-0.02, 0.02]$ + ref.\\
Bike Base Position $Z$            & m   & $\mathcal{U}[0.0, 0.1]$    & $0.15$ (Fixed)   & $\mathcal{U}[0.05, 0.15]$  & $\mathcal{U}[0.0, 0.05]$   & $\mathcal{U}[0.0, 0.02]$ + ref.\\
Bike Base Orientation Roll        & rad & $\mathcal{U}[-0.1, 0.1]$   & $\pi$ (Fixed)    & $\mathcal{U}[-0.1, 0.1]$   & $\mathcal{U}[-0.1, 0.1]$   & $\mathcal{U}[-0.1, 0.1]$ + ref.\\
Bike Base Orientation Pitch       & rad & $\mathcal{U}[-0.05, 0.05]$ & $-\pi/6$ (Fixed) & $\mathcal{U}[-0.05, 0.05]$ & $\mathcal{U}[-0.02, 0.02]$ & $\mathcal{U}[-0.05, 0.05]$ + ref.\\
Bike Base Orientation Yaw         & rad & $\mathcal{U}[-0.1, 0.1]$   & $0$ (Fixed)      & $\mathcal{U}[-\pi, \pi]$   & $\mathcal{U}[-0.1, 0.1]$   & $\mathcal{U}[-0.1, 0.1]$ + ref.\\
Bike Base Linear Velocity $X$     & m/s & $\mathcal{U}[-0.1, 0.1]$   & $-$              & $\mathcal{U}[-0.1, 0.1]$   & $\mathcal{U}[-0.1, 0.1]$   & $\mathcal{U}[-0.1, 0.1]$ + ref.\\
Bike Base Linear Velocity $Y$     & m/s & $-$ & $-$ & $-$ & $-$   &  $\mathcal{U}[-0.1, 0.1]$ + ref. \\
Bike Base Linear Velocity $Z$     & m/s & $-$ & $-$ & $-$ & $-$   &  $\mathcal{U}[-0.1, 0.1]$ + ref. \\
Bike Base Angular Velocity Roll   & rad/s & $-$ & $-$ & $-$ & $-$ &  $\mathcal{U}[-0.1, 0.1]$ + ref. \\
Bike Base Angular Velocity Pitch  & rad/s & $-$ & $-$ & $-$ & $-$ &  $\mathcal{U}[-0.05, 0.05]$ + ref. \\
Bike Base Angular Velocity Yaw    & rad/s & $-$ & $-$ & $-$ & $-$ &  $\mathcal{U}[-0.1, 0.1]$ + ref. \\
\midrule
\ourrow\multicolumn{7}{l}{\textbf{(C) Initial Joint State Randomization$^2$}} \\
Neck Joint $q_n$                        & rad   & $\mathcal{U}[0.0, 0.05]$   & $0.0$ (Fixed)            & $\mathcal{U}[0.0, 1.2]$  & $\mathcal{U}[0.0, 0.05]$   &  $\mathcal{U}[0.05, 0.2]$ or $\mathcal{U}[-0.1, 0.1]$ + ref.\\
Head Joint $q_h$                        & rad   & $\mathcal{U}[-0.05, 0.00]$ & $0.0$ (Fixed)            & $\mathcal{U}[-1.2, 0.0]$ & $\mathcal{U}[-0.05, 0.00]$ &  $\mathcal{U}[-0.2, -0.05]$ or $\mathcal{U}[-0.1, 0.1]$ + ref.\\
Steering Joint $q_s$                    & rad   & $\mathcal{U}[-0.5, 0.5]$   & $\mathcal{U}[-0.5, 0.5]$ & $\mathcal{U}[-0.5, 0.5]$ & $\mathcal{U}[-0.5, 0.5]$   &  $\mathcal{U}[-0.1, 0.1]$ or $\mathcal{U}[-0.1, 0.1]$ + ref.\\
Head, Neck, and Steering Joint Velocity & rad/s & $-$ & $-$ & $-$ & $-$ & $0.0$ (fixed) or $\mathcal{U}[-0.1, 0.1]$ + ref. \\
\midrule
\ourrow\multicolumn{7}{l}{\textbf{(D) External Disturbance Forces (Push Robots)}} \\
Body Velocity Disturbance $X$ & m/s & $\mathcal{U}[-0.7, 0.7]$ & $\mathcal{U}[-0.7, 0.7]$ & $\mathcal{U}[-0.7, 0.7]$ & $\mathcal{U}[-0.7, 0.7]$ & $\mathcal{U}[-0.1, 0.1]$ \\
Body Velocity Disturbance $Y$ & m/s & $\mathcal{U}[-0.2, 0.2]$ & $\mathcal{U}[-0.1, 0.1]$ & $\mathcal{U}[-0.2, 0.2]$ & $\mathcal{U}[-0.2, 0.2]$ & $\mathcal{U}[-0.1, 0.1]$ \\
Body Velocity Disturbance $Z$ & m/s & $-$ & $\mathcal{U}[-0.1, 0.1]$ & $-$ & $-$ & $\mathcal{U}[-0.1, 0.1]$ \\
\midrule
\ourrow\multicolumn{7}{l}{\textbf{(E) Heightmap \& Obstacle Elements}} \\
Heightmap Drift Range & m & $\mathcal{U}[-0.1, 0.1]$ & $-$ & $-$ & $-$ & $-$ \\
Heightmap Bias Range & m & $\mathcal{U}[-0.05, 0.05]$ & $-$ & $-$ & $-$ & $-$ \\
Obstacle Table Height & m & $\mathcal{U}[0.0, 1.0]$ & $-$ & $-$ & $-$ & $-$ \\
Obstacle Table Width & m & $\mathcal{U}[1.0, 3.0]$ & $-$ & $-$ & $-$ & $-$ \\
Obstacle Flat Ratio & ratio & $0.1$ (Fixed) & $-$ & $-$ & $-$ & $-$ \\
Obstacle Gap Length & m & $\mathcal{U}[0.0, 1.0]$ & $-$ & $-$ & $-$ & $-$ \\
\midrule
\ourrow\multicolumn{7}{l}{\textbf{(F) Reference State Initialization (RSI)}} \\
Phase Initialization Range          & ratio & $-$ & $-$ & $-$ & $-$ & $\mathcal{U}[-0.9, 0.9]$ \\
Absolute Initialization Ratio       & ratio & $-$ & $-$ & $-$ & $-$ & $0.5$ (Fixed) \\
\bottomrule
\multicolumn{7}{l}{\small$^1$The base states of the imitation stunts are initialized and randomized with respect to the reference.} \\
\multicolumn{7}{l}{\small$^2$The joint states of the imitation stunts are initialized and randomized either from scratch or with respect to the reference.} \\
\end{tabular}
}
\end{table*}

\begin{table*}[t]
\centering
\caption{\bf Termination Criteria}
\label{table_terminations_all}  
\begin{tabular}{l ccccc}
\toprule
\textbf{\shortstack[l]{Termination Term\\~}} & \textbf{\shortstack{Waypoint\\Following}} & \textbf{\shortstack{Pose\\Reaching}} & \textbf{\shortstack{SE2 Twist\\Tracking}} & \textbf{\shortstack{Guided\\Tracking}} & \textbf{\shortstack{Motion\\Imitation}} \\
\midrule
\ourrow\multicolumn{6}{l}{\textbf{(a) Kinematic \& Joint Boundary Limits}} \\
Steering Joint Position Limit & \checkmark & \checkmark & \checkmark & - & \checkmark \\
Steering Joint Velocity Limit & - & \checkmark & \checkmark & - & \checkmark \\
Wheel Joint Velocity Limit    & \checkmark & \checkmark & \checkmark & - & \checkmark \\
Neck Joint Position Limit     & \checkmark & \checkmark & \checkmark & \checkmark & \checkmark \\
Head Joint Position Limit     & \checkmark & \checkmark & - & \checkmark & \checkmark \\
Fin Joints Position Limit     & \checkmark & \checkmark & - & - & - \\
\midrule
\ourrow\multicolumn{6}{l}{\textbf{(b) Ground Collisions \& Unsafe Orientation Excursions}} \\
Ground Collision (Head Only)                   & \checkmark & - & - & - & - \\
Ground Collision (Base \& Neck)                & - & \checkmark & - & - & - \\
Ground Collision (Base, Head \& Neck)          & - & - & \checkmark & - & \checkmark \\
Base Angular Velocity / Orientation Rate Limit & \checkmark & \checkmark & - & - & - \\
Inverted Roll Orientation Limit                & - & \checkmark & - & - & - \\
\midrule
\ourrow\multicolumn{6}{l}{\textbf{(c) Actuator Effort, Power \& Dynamic Thresholds}} \\
Motor Effort Limit                & \checkmark & \checkmark & - & \checkmark & \checkmark \\
Motor Velocity Limit              & - & \checkmark & - & - & - \\
Neck and Head Joint Torque Limits & - & \checkmark & - & - & - \\
Mechanical Joint Power Limit      & - & - & - & - & \checkmark \\
\midrule
\ourrow\multicolumn{6}{l}{\textbf{(d) Trajectory Deviations \& Contact Drift Boundaries}} \\
High Contact Velocity Limit          & \checkmark & \checkmark & - & \checkmark & \checkmark \\
Guideline Trajectory Deviation             & - & - & - & \checkmark & - \\
Orientation Keyframe Deviation             & - & - & - & \checkmark & - \\
Reference Trajectory Position Deviation    & - & - & - & - & \checkmark \\
Reference Trajectory Orientation Deviation & - & - & - & - & \checkmark \\
\bottomrule
\end{tabular}
\end{table*}

\begin{table*}[t]
\centering
\caption{\bf Curriculum Learning}
\label{table_curr}  
\begin{tabular}{l cccc}
\toprule
\textbf{\shortstack[l]{Curriculum Term\\~}} & \textbf{\shortstack{Waypoint\\Following}} & \textbf{\shortstack{Pose\\Reaching}} &  \textbf{\shortstack{Guided\\Tracking}} & \textbf{\shortstack{Motion\\Imitation}} \\
\midrule
Joint Position Limits             & after 5000 iter. &  after 1000 iter.             & 2000 iter. after success & --\\
Orientation Rate Limit            & after 5000 iter. &  --                           & --                       & --\\
High Contact Velocity Limit       & after 6000 iter. &  --                           & 500 iter. after success  & after 5000 iter.\\
Motor Effort Limit                & --               &  linear interp. 5500 to 6500  & 3000 iter. after success & after 1000 iter.\\
Motor Velocity Limit              & --               &  linear interp. 4000 to 5000  & --                       & --\\
Neck and Head Joint Torque Limits & --               &  linear interp. 6500 to 7500  & --                       & --\\
Mechanical Joint Power Limit      & --               &  --                           & --                       & after 5000 iter.\\
\bottomrule
\end{tabular}
\end{table*}

\begin{table*}[t]
\centering
\caption{\bf Common Training Algorithm Params}
\label{table_common_hyper_params}  
\begin{tabular}{l c}
\toprule
\textbf{Parameter} & \textbf{Value} \\
\midrule
Num. Robots                           & 4096 \\
Num. Steps Per Interval               & 24 \\
Actor/Critic Init. Noise Std          & value \\
Actor/Critic Network Activation Func. & ELU \\
Value Loss Coefficient                & 1 \\
Clip Parameter                        & 0.2 \\
Num. Learning Epochs                  & 5 \\
Num. Mini Batches                     & 4 \\
Learning Rate                         & $1\times10^{-3}$ \\
Schedule                              & Adaptive \\
Discount Factor~$\gamma$              & 0.99 \\
GAE $\lambda$                         & 0.95 \\
Desired KL                            & 0.01 \\
Max. Grad Norm                        & 1.0 \\
\bottomrule
\end{tabular}
\end{table*}

\begin{table*}[t]
\centering
\caption{\bf Task Specific Training Algorithm Params}
\label{table_other_hyper_params}  
\begin{tabular}{l ccccc}
\toprule
\textbf{\shortstack[l]{Parameter\\~}} & \textbf{\shortstack{Waypoint\\Following}} & \textbf{\shortstack{Pose\\Reaching}} & \textbf{\shortstack{SE2 Twist\\Tracking}} & \textbf{\shortstack{Guided\\Tracking}} & \textbf{\shortstack{Motion\\Imitation}} \\
\midrule
Iteration Number                 & 30000                   & 30000                      & 25000                   & 25000               & 15000 \\
PPO Entropy Coefficient          & 0.006                   & 0.0035                     & 0.0035                  & 0.002               & 0.0025 \\
Actor/Critic Network Hidden Dim. & [256, 128, 128, 64, 32] & \multicolumn{2}{c}{[256, 128, 64, 32, 16]}   &  \multicolumn{2}{c}{[512, 256, 128]} \\
\bottomrule
\end{tabular}
\end{table*}

\end{document}